\documentclass[lettersize,journal]{IEEEtran}
\usepackage{makecell}
\usepackage{booktabs}
\usepackage{multirow}
\usepackage{bbding}
\usepackage{subfigure}
\usepackage{array}
\usepackage[caption=false,font=normalsize,labelfont=sf,textfont=sf]{subfig}
\usepackage{textcomp}
\usepackage{stfloats}
\usepackage{url}
\usepackage{verbatim}
\usepackage{graphicx}
\usepackage{cite}
\usepackage{balance}
\usepackage{amsmath,amsfonts}
\usepackage{algorithm}
\usepackage{algpseudocode}

\begin{document}

\title{A Minimal Set of Parameters Based Depth-Dependent Distortion Model and Its Calibration Method for Stereo Vision Systems}
\author{Xin Ma, Puchen Zhu, Xiao Li, Xiaoyin Zheng, Jianshu Zhou,~\IEEEmembership{Member,~IEEE,} Xuchen Wang, Kwok Wai Samuel Au,~\IEEEmembership{Member,~IEEE}}

\maketitle

\begin{abstract}
Depth position highly affects lens distortion, especially in close-range photography, which limits the measurement accuracy of existing stereo vision systems. Moreover, traditional depth-dependent distortion models and their calibration methods have remained complicated. In this work, we propose a minimal set of parameters based depth-dependent distortion model (MDM), which considers the radial and decentering distortions of the lens to improve the accuracy of stereo vision systems and simplify their calibration process. In addition, we present an easy and flexible calibration method for the MDM of stereo vision systems with a commonly used planar pattern, which requires cameras to observe the planar pattern in different orientations. The proposed technique is easy to use and flexible compared with classical calibration techniques for depth-dependent distortion models in which the lens must be perpendicular to the planar pattern. The experimental validation of the MDM and its calibration method showed that the MDM improved the calibration accuracy by 56.55\% and 74.15\% compared with the Li's distortion model and traditional Brown's distortion model. Besides, an iteration-based reconstruction method is proposed to iteratively estimate the depth information in the MDM during three-dimensional reconstruction. The results showed that the accuracy of the iteration-based reconstruction method was improved by 9.08\% compared with that of the non-iteration reconstruction method.
\end{abstract}


\section{Introduction}
\IEEEPARstart{L}{ens} distortion existing in optic measurement, which is usually caused by refraction and focusing of light rays through camera lenses, will result in changes in the shape and size of an image, subsequently affecting the image's quality and accuracy. The lens distortion is mainly classified as a radial distortion and a decentering (or tangential[1]) distortion[2]. Radial distortion occurs when straight lines in the image appear curved or bulging at edges, causing a distortion in the proportion between the center and edges of the image. On the other hand, the decentering distortion refers to bending or warping of straight lines in the image along the horizontal or vertical direction, resulting in an overall skewed or distorted shape. The occurrence of lens distortion is due to the construction of the lens and the propagation of light rays within it. In the design and manufacturing process of lenses, certain unavoidable flaws exist, leading to inaccurate refraction and focusing of light rays, thereby causing distortions.

Accurate lens distortion calibration is the basis of precise vision measurement, especially in stereo vision systems [3]. Conventional lens distortion models, which consider radial and decentering lens distortions [4], [5] usually take the spatial constraints of the calibration target as the target and employ the nonlinear least-square optimization method to conduct the calibration (Zhang's [6] method). However, these models neglect the effect of depth information on lens distortion [7], [8]. When measuring a distant target, the performance of these models is satisfactory. However, for large depths in close-range photogrammetry, depth information significantly impacts lens distortion; thus, adopting these models [9]-[11] can result in the low accuracy of the vision measurement system.

Magill [12] explained the relationship between radial distortion and magnification for the first time. However, the unknown parameters (distortion function for inverted infinite focus) in the distortion model could not be calibrated. Then, Brown presented a more applicable depth-dependent radial distortion model [13]. Consequently, Fryer proposed a depth-dependent decentering distortion model [14]. In these depth-dependent distortion models, no distortion function exists for inverted infinite focus, making the calibration of the lens distortion model possible. However, distortion functions at different focused depths must be calibrated, which entails adjusting the lens to focus on planes at different depths during the calibration process. To reduce the complexity of the calibration process, Alvarez derived a new radial distortion model by combining the Brown and Fraser [15] models. During the calibration, the lens focus does not need to be adjusted, increasing the stability of the calibration [16]. However, the decentering and in-plane affine distortion parameters are not calibrated. Consequently, Sun [17] and Li [18] proposed two depth-dependent distortion models, which consider both decentering and radial lens distortions. 
However, these depth-dependent lens distortion models have more than 16 unknown parameters, which makes the depth-dependent lens distortion models complicated. For example, the depth-dependent radial lens distortion model and decentering lens distortion model in [18] have 9 unknown parameters and 8 unknown parameters respectively, which need to be calibrated separately in 9 steps in total. The above-mentioned lens distortion models are listed in Table I. During the calibration procedures, especially when estimating the radial and decentering lens distortion parameters in the defocused plane in the depth-dependent lens distortion model, the checkerboard orientations need to be adjusted by mounting the checkerboard on an electric control platform and the camera needs to be adjusted to be perpendicular to the checkerboard. Thus, it will take plenty of time to adjust the checkerboard orientations during the calibration procedures and the additional equipment such as the electric control platform contribute to an increase in costs, which makes the calibration procedures inefficient. Besides, no effective three-dimensional (3D) reconstruction method is proposed in the reference for handling the input-output coupling (depth information) issue.

\begin{table*}[]
\centering
\caption{Comparison among different lens distortion models.}
\begin{tabular}{ccccccc}
\hline
\multirow{3}{*}{Method}                                           & \multirow{3}{*}{\begin{tabular}[c]{@{}c@{}}Number \\ of\\ Parameters\end{tabular}} & \multirow{3}{*}{\begin{tabular}[c]{@{}c@{}}Considering\\ Depth-\\ dependent\end{tabular}} & \multirow{3}{*}{\begin{tabular}[c]{@{}c@{}}Considering\\ Radial \\ Distortion\end{tabular}} & \multirow{3}{*}{\begin{tabular}[c]{@{}c@{}}Considering\\ Decentering \\ Distortion\end{tabular}} & \multirow{3}{*}{Advantages}                                                                                              & \multirow{3}{*}{Disadvantages}                                                                                             \\
                                                                  &                                                                                    &                                                                                           &                                                                                             &                                                                                                  &                                                                                                                          &                                                                                                                            \\
                                                                  &                                                                                    &                                                                                           &                                                                                             &                                                                                                  &                                                                                                                          &                                                                                                                            \\ \hline
\begin{tabular}[c]{@{}c@{}}Brown's \\ Model {[}13{]}\end{tabular} & 6                                                                                  & \Checkmark                                                                                       & \Checkmark                                                                                         & \Checkmark                                                                                              & \begin{tabular}[c]{@{}c@{}}The foundation of\\ other distortion models\end{tabular}                                      & \begin{tabular}[c]{@{}c@{}}No distrotion functions exists\\ for inverted infinite focus\end{tabular}                       \\
\multirow{2}{*}{RFDM {[}7{]}}                                     & \multirow{2}{*}{23}                                                                & \multirow{2}{*}{\XSolidBrush}                                                                       & \multirow{2}{*}{\Checkmark}                                                                        & \multirow{2}{*}{\Checkmark}                                                                             & \multirow{2}{*}{\begin{tabular}[c]{@{}c@{}}Accurate and flexible in\\ 3D shape measurement\end{tabular}}                 & \multirow{2}{*}{\begin{tabular}[c]{@{}c@{}}Depth information is not\\ considered compared with {[}9{]}\end{tabular}}       \\
                                                                  &                                                                                    &                                                                                           &                                                                                             &                                                                                                  &                                                                                                                          &                                                                                                                            \\
\multirow{2}{*}{DRDM {[}9{]}}                                     & \multirow{2}{*}{20}                                                                & \multirow{2}{*}{\Checkmark}                                                                      & \multirow{2}{*}{\Checkmark}                                                                        & \multirow{2}{*}{\Checkmark}                                                                             & \multirow{2}{*}{\begin{tabular}[c]{@{}c@{}}Accurate and flexible\\ under large-field-of-view\end{tabular}}               & \multirow{2}{*}{\begin{tabular}[c]{@{}c@{}}More parameters to be \\ calibrated compared with {[}16{]}\end{tabular}}        \\
                                                                  &                                                                                    &                                                                                           &                                                                                             &                                                                                                  &                                                                                                                          &                                                                                                                            \\
\multirow{2}{*}{ADDM {[}16{]}}                                    & \multirow{2}{*}{16}                                                                & \multirow{2}{*}{\Checkmark}                                                                      & \multirow{2}{*}{\Checkmark}                                                                        & \multirow{2}{*}{\XSolidBrush}                                                                              & \multirow{2}{*}{Valid in any scenario}                                                                                   & \multirow{2}{*}{\begin{tabular}[c]{@{}c@{}}Decentering distortion is not\\ considered compared with {[}18{]}\end{tabular}} \\
                                                                  &                                                                                    &                                                                                           &                                                                                             &                                                                                                  &                                                                                                                          &                                                                                                                            \\
\multirow{2}{*}{Li's Model {[}18{]}}                              & \multirow{2}{*}{16}                                                                & \multirow{2}{*}{\Checkmark}                                                                      & \multirow{2}{*}{\Checkmark}                                                                        & \multirow{2}{*}{\Checkmark}                                                                             & \multirow{2}{*}{\begin{tabular}[c]{@{}c@{}}Avoiding manually \\ adjusting of the focus\end{tabular}}                     & \multirow{2}{*}{\begin{tabular}[c]{@{}c@{}}More parameters to be\\ calibrated compared with MDM\end{tabular}}              \\
                                                                  &                                                                                    &                                                                                           &                                                                                             &                                                                                                  &                                                                                                                          &                                                                                                                            \\
\multirow{2}{*}{MDM}                                              & \multirow{2}{*}{8}                                                                 & \multirow{2}{*}{\Checkmark}                                                                      & \multirow{2}{*}{\Checkmark}                                                                        & \multirow{2}{*}{\Checkmark}                                                                             & \multirow{2}{*}{\begin{tabular}[c]{@{}c@{}}The camera is not required\\ to be perpendicular to the pattern\end{tabular}} & \multirow{2}{*}{\begin{tabular}[c]{@{}c@{}}Hard to find accurate \\ optimized initial values\end{tabular}}                 \\
                                                                  &                                                                                    &                                                                                           &                                                                                             &                                                                                                  &                                                                                                                          &                                                                                                                            \\ \hline
\end{tabular}
\label{table 1}
\end{table*}

In this paper, we propose a minimal set of parameters based depth-dependent lens distortion model (MDM), which considers the radial and decentering distortions of the lens to improve the accuracy of stereo vision systems and simplify their calibration process. Compared with [18], the minimal set of parameters in the MDM has a total of 8 unknown parameters of both radial and decentering lens distortion models, which makes the model more simplified. Besides, we present an easy and flexible calibration method for the MDM of stereo vision systems with a commonly used planar pattern, which requires cameras to observe the planar pattern at different orientations. Compared with some general calibration methods for the depth-dependent distortion models, such as the model proposed by Li and Sun, the proposed technique does not need the camera to be adjusted to be perpendicular to the planar pattern during the calibration, which is easy to use and flexible.  Several experiments have been conducted to validate the MDM and its calibration method. The results showed the MDM improves the calibration accuracy by 56.55\% and 74.15\% compared with the Li's distortion model and conventional Brown's distortion model. Moreover, an iteration-based reconstruction method is proposed to iteratively estimate the depth information in the MDM during 3D reconstruction for handling the input-output coupling (depth information) issue. The results showed that the iteration-based reconstruction method improved the calibration accuracy by 9.08\% compared with the non-iteration reconstruction method.

This paper is organized as follows: Section II introduces the minimal set of parameters based depth-dependent lens distortion model. Section III introduces the calibration method for the MDM and elaborates on the 3D reconstruction method for the MDM. Section IV provides the experimental details, and Section V outlines the conclusions of the paper.

\section{Minimal Set of Parameters Based Depth-Dependent Lens Distortion Model}
In this section, the minimal set of parameters based depth-dependent radial lens distortion model (MDM-R) and the minimal set of parameters based depth-dependent decentering lens distortion model (MDM-D) are proposed. 
\vspace{-2 mm} 
\subsection{Minimal Set of Parameters Based Depth-Dependent Radial Lens Distortion Model}
In this section, we simplified the classical depth-dependent radial lens distortion model based on the Magill model.
The Magill model was first simplified by Brown in 1971 [13], which overcame the limited practicality. The simplified model is shown in (1):

\begin{equation}
\label{e1}
\left\{\begin{array}{l}
\delta r_s=\alpha_s \cdot \delta r_{s_m}+\left(1-\alpha_s\right) \cdot \delta r_{s_k} \\
\alpha_s=\frac{\left(s_k-s_m\right) \cdot(s-f)}{\left(s_k-s\right) \cdot\left(s_m-f\right)}
\end{array}\right.
\end{equation}
where $f$ denotes the focal length. When the lens is adjusted to focus on the distances $s_m$ and $s_k$, the radial distortions in the focal planes are defined as $\delta r_{s_m}$  and $\delta r_{s_k}$. Then, the distortion function $\delta r_s$  for the lens adjusted to focus on distance $s$ can be computed. The $I$-th radial distortion coefficients $K_I^s$ for the focused object plane at distance $s$ are described by:

\begin{equation}
\label{e2}
K_I^s=\alpha_s \cdot K_I^{s_m}+\left(1-\alpha_s\right) \cdot K_I^{s_k} 
\end{equation}
where $I=1,2.$, and $K_I^{s_m}$ and $K_I^{s_k}$ represent the $I$-th radial distortion coefficients when the lens is adjusted to focus on the distances $s_m$ and $s_k$, respectively. However, an empirical model was proposed by Fraser and Shortis for the distortion of any defocused plane [15]:

\begin{equation}
\label{e3}
K^{s, s_p}=K^s+\varepsilon \cdot\left(K^{s_p}-K^s\right)
\end{equation}
where $K^{s, s_p}$ represents the radial distortion coefficient at defocused distance $s_p$ for the lens focused at distance $s$; $\varepsilon$ is a constant correction factor; $K^{s_p}$ and $K^{s}$ denote the radial distortion coefficients in the focal planes at distances $s_p$ and $s$, respectively. According to (3), the radial distortion function $\delta^{s, s_n}$ at $s_n$ can be obtained when the lens is adjusted to focus at distance $s$ :

\begin{equation}
\label{e4}
\left\{\begin{array}{c}
\delta r^{s, s_m}=\delta r^s+\varepsilon \cdot\left(\delta r^{s_m}-\delta r^s\right) \\
\!\delta r^{s, s_n}-\delta r^{s, s_m}=\varepsilon \! \cdot\!\left(\delta r^{s_n}-\delta r^s\right)\!-\!g\! \cdot\!\left(\delta r^{s_m}-\delta r^s\right)
\end{array}\right.
\end{equation}

Thus, by adopting the results to the distortion of a point in the defocused plane at distance $s_k$, the relationship among $\delta r^{s, s_n}$, $\delta r^{s, s_m}$, and $\delta r^{s, s_k}$ can be described as:

\begin{equation}
\label{e5}
\left\{\begin{array}{l}
\delta r^{s, s_n}=\delta r^{s, s_m}+\frac{\left(s_m-s_n\right)}{\left(s_m-s\right)}\!\!\cdot\!\!\frac{(s-f)}{\left(s_n-f\right)} \!\!\cdot\!\!\left(\delta r^s-\delta r^{s, s_m}\right) \\
\delta r^{s, s_n}=\delta r^{s, s_k}+\frac{\left(s_k-s_n\right)}{\left(s_k-s\right)} \!\!\cdot\!\! \frac{(s-f)}{\left(s_n-f\right)} \!\!\cdot\!\!\left(\delta r^s-\delta r^{s, s_k}\right) \\
\delta r^{s, s_m}=\delta r^{s, s_k}+\frac{\left(s_k-s_m\right)}{\left(s_k-s\right)} \!\!\cdot\!\! \frac{(s-f)}{\left(s_n-f\right)} \!\!\cdot\!\!\left(\delta r^s-\delta r^{s, s_k}\right)
\end{array}\right.
\end{equation}
Then (6) can be obtained by eliminating the focus distance, and the distortion in the focal plane:

\begin{equation}
\label{e6}
\left\{\begin{array}{l}
\delta r^{s, s_m}=\delta r^{s, s_k}+\frac{s_k-s_m}{s_k-s} \cdot \frac{s-f}{s_m-f} \cdot\left(\delta r^s-\delta r^{s, s_k}\right) \\
\delta r^{s, s_n}=\delta r^{s, s_k}+\frac{s_k-s_n}{s_k-s} \cdot \frac{s-f}{s_n-f} \cdot\left(\delta r^s-\delta r^{s, s_k}\right)
\end{array}\right.
\end{equation}

According to (6), $\delta r^{s, s_n}$  can be written as 

\begin{equation}
\label{e7}
\!\delta \!r^{\!s,\! s_n\!}\!=\!\frac{\delta r^{\!s, \!s_k\!} \!\!\cdot\!\!\left(\!s_m\!\!-\!\!f\!\right)\! \!\cdot\!\!\left(\!s_k\!-\!s_m\!\!\right)\!\!+\!\!\left(\!s_k\!\!-\!\!s_n\!\!\right) \!\!\cdot\!\!\left(\!\!s_k\!\!-\!\!f\!\right)\! \!\cdot\!\!\left(\delta r^{\!s\!, \!s_m\!}\!\!-\!\!\delta r^{\!s\!, \!s_k\!}\right)}{\left(\!s_m\!-\!f\!\right) \!\cdot\!\left(\!s_k\!-\!s_m\!\right)}
\end{equation}

When the lens is adjusted to focus at distance $s$, the radial distortion coefficients in any defocused plane with the depth of $s_n$ can be obtained via two distortions corresponding to object distances $s_m$ and $s_k$, respectively [18]:

\begin{equation}
\label{e8}
\!K_I^{\!s\!,\! s_n\!}\!=\!\frac{\!K_I^{\!s\!, \!s_k\!} \!\!\cdot\!\!\!\left(\!s_m\!\!\!-\!\!\!f\!\right)\!\! \cdot\!\!\left(\!s_k\!\!-\!\!s_m\!\right)\!\!+\!\!\left(\!s_k\!\!\!-\!\!\!s_n\!\right)\! \!\cdot\!\!\left(\!s_k\!-\!f\!\right)\! \!\cdot\!\!\left(\!K_I^{\!s\!,\! s_m\!}\!\!-\!\!K_I^{\!s\!,\! s_k\!}\right)}{\left(\!s_m\!-\!f\!\right) \!\cdot\!\left(\!s_k\!-\!s_m\!\right)} 
\end{equation}
According to (8), $K_I^{s, s_m}$, $K_I^{s, s_k}$, $s_m$, $s_k$, and $f$ are known since two object planes are intentionally set. Thus, $K_I^{s, s_n}$ in (8) depends on $s_n$ only, and (8) can be further simplified as follows:

\begin{equation}
\label{e9}
\!\!\left\{\begin{aligned}
\delta r_u^R & \!=\!\bar{u} \!\!\cdot\!\! \left[1\!\!+\!\!\left(A_1\!\!-\!\!B_1 \cdot s_n\right) \cdot r^2\!\!+\!\!\left(A_2\!\!-\!\!B_2 \cdot s_n\right) \cdot r^4\!\!+\!\!\ldots\right] \\
& \!=\!\bar{u} \!\!\cdot\!\! \left[1\!\!+\!\!K_1^{s, s_n} \cdot r^2\!\!+\!\!K_2^{s, s_n} \cdot r^4+\ldots\right] \\
\delta r_v^R & \!=\!\bar{v} \!\!\cdot\!\! \left[1\!\!+\!\!\left(A_1\!\!-\!\!B_1 \cdot s_n\right) \cdot r^2\!\!+\!\!\left(A_2\!\!-\!\!B_2 \cdot s_n\right) \cdot r^4\!\!+\!\!\ldots\right] \\
& \!=\!\bar{v} \!\!\cdot\!\! \left[1+K_1^{s, s_n} \cdot r^2+K_2^{s, s_n} \cdot r^4+\ldots\right] \\
K_I^{s\!,\! s_n} & =A_I-B_I \cdot s_n \\
A_I & \!=\!\frac{K_I^{s, s_k} \!\!\cdot\!\! \left(\!s_m\!\!-\!\!f\!\right) \!\!\cdot\!\! \left(\!s_k\!\!-\!\!s_m\!\right)\!\!+\!\!s_k \!\!\cdot\!\! \left(\!s_k\!\!-\!\!f\!\right) \!\!\cdot\!\! \left(\!K_I^{s, s_m}\!\!-\!\!K_I^{s, s_k}\!\right)}{\left(s_m-f\right) \!\!\cdot\!\! \left(s_k-s_m\right)} \\
B_I & \!=\!\frac{\left(s_k-f\right) \!\!\cdot\!\! \left(K_I^{s, s_m}-K_I^{s, s_k}\right)}{\left(s_m-f\right) \!\!\cdot\!\! \left(s_k-s_m\right)}
\end{aligned}\right.
\end{equation} 

Here $A_I$ and $B_I$ are the $I$-th simplified radial lens distortion parameters in the short-distance defocused plane with the depth of $s_n$; $\delta r_u^R$ and $\delta r_v^R$ are the radial distortion functions in the $u$ and $v$ directions, respectively;  $(\bar{u}, \bar{v})$ is the uncorrected corner point image coordinate in the checkerboard image; $\left(u_0, v_0\right)$ is the principal point coordinate; $r=\sqrt{\left(\bar{u}-u_0\right)^2+\left(\bar{v}-v_0\right)^2}$ is the distortion radius of the image point. Then, the MDM-R can be expressed as shown in (9).

\subsection{Minimal Set of Parameters Based Depth-Dependent Decentering Lens Distortion Model}
In this section, we simplified the classical decentering lens distortion model based on the Brown's distortion model. The decentering lens distortion model proposed by Brown in 1966 [19] is described as follows:

\begin{equation}
\label{e10}
\left\{\begin{array}{c}
\delta r_u\!=\!\left(1-\frac{f}{s}\right) \!\!\cdot\!\!\left[P_1 \cdot\left(r^2+2 \cdot \bar{u}^2\right)+2 \cdot P_2 \cdot \bar{u} \cdot \bar{v}\right] \\
\delta r_v\!=\!\left(1-\frac{f}{s}\right) \!\!\cdot\!\!\left[P_2 \cdot\left(r^2+2 \cdot \bar{v}^2\right)+2 \cdot P_1 \cdot \bar{u} \cdot\bar{v}\right]\\
r^{s,s^{\prime}}=\frac{s-f}{s^{\prime}-f} \cdot \frac{s^{\prime}}{s}
\end{array}\right.
\end{equation}
where $\left(1-\frac{f}{s}\right) \cdot r^{s, s^{\prime}}$  denotes the compensation coefficient; $\delta r_u$ and $\delta r_v$ represent the components of the decentering distortion in the $u$ and $v$ directions, respectively; $s$ and $s^{\prime}$  are the object distances corresponding to the two focal planes, respectively.

Based on (10), a depth-dependent decentering distortion model is proposed in [18], which is described as follows:

\begin{equation}
\label{e13}
P_I^{\!s\!, \!s_n\!}=\frac{P_I^{\!s\!,\! s_k\!} \!\cdot\!\left(1\!-\!s_m^2\right)\!+\!P_I^{\!s\!,\! s_m\!}\left(\!s_m\! \cdot\! s_k\!-\!1\!\right)}{P_I^{\!s\!, \!s_k\!} \!\cdot\!\left(1\!-\!s_m\! \cdot\! s_n\right)\!+\!P_I^{s\!, \!s_m\!}\left(s_n \!\cdot\! s_k\!-\!1\right)} \!\cdot\! \frac{s_n}{s_m}\! \cdot\! P_I^{s\!, \!s_m\!}  
\end{equation}
where $P_I^{s, s_m}$, $P_I^{s, s_k}$, and $P_I^{s, s_n}$ represent the $I$-th decentering distortion parameters in the defocused planes at the object distances $s_m$, $s_k$, and $s_n$ when the lens is adjusted to focus at distance $s$, respectively. According to (13), the decentering distortion parameters $P_I^{s, s_n}$ in any defocused plane depend on only the object distance $s_n$ since $P_I^{s, s_m}$, $P_I^{s, s_k}$, $s_m$ and $s_k$ are known. However, (13) also has 6 decentering lens distortion parameters to be calibrated separately, which makes the decentering lens distortion model complicated. Similarly, the checkerboard orientations need to be adjusted by mounting the checkerboard on an electric control platform, and the camera also needs to be adjusted to be perpendicular to the checkerboard to estimate $P_I^{s, s_m}$ and $P_I^{s, s_k}$ during the calibration procedures, which makes the calibration procedures inefficient.
Thus, a minimal set of decentering distortion parameters is described, and the decentering distortion model can be further simplified as follows:

\begin{equation}
\label{e14}
\left\{\begin{aligned}
\delta r_u^D & =\frac{1}{C_1+\frac{D_1}{s_n}} \!\cdot\!\left(r^2+2 \!\cdot\! \bar{u}^2\right)+2 \!\cdot\! \frac{1}{C_2+\frac{D_2}{s_n}} \!\cdot\! \bar{u} \!\cdot\! \bar{v} \\
& =P_1 \!\cdot\!\left(r^2+2 \!\cdot\! \bar{u}^2\right)+2 \!\cdot\! P_2 \!\cdot\! \bar{u} \!\cdot\! \bar{v} \\
\delta r_v^D & =\frac{1}{C_2+\frac{D_2}{s_n}} \!\cdot\!\left(r^2+2 \!\cdot\! \bar{v}^2\right)+2 \!\cdot\! \frac{1}{C_1+\frac{D_1}{s_n}} \!\cdot\! \bar{u} \!\cdot\! \bar{v} \\
& =P_2 \!\cdot\!\left(r^2+2 \!\cdot\! \bar{v}^2\right)+2 \!\cdot\! P_1 \!\cdot\! \bar{u} \!\cdot\! \bar{v} \\
P_I^{s, s_n} & =\frac{1}{C_I+\frac{D_I}{s_n}}
\end{aligned}\right.
\end{equation}

Here $C_I$ and $D_I$ are the $I$-th decentering lens distortion parameters in the short-distance defocused plane with the depth of $s_n$; $\delta r_u^D$ and $\delta r_v^D$ are the decentering distortion functions in the $u$ and $v$ directions, respectively. The MDM-D is described as shown in (14), which has a minimal set of 4 dencentering lens distortion parameters.

\section{Calibration and 3D Reconstruction Methods for the MDM}
In this section, a flexible three-step calibration method for the MDM is proposed. First, the intrinsic and extrinsic camera parameters of the stereo vision system are calculated by the Zhang's calibration method. Then, the MDM is estimated using linear constraints. Finally, the intrinsic and extrinsic camera parameters of the stereo vision system and the parameters in the MDM are optimized based on the distance constraints between adjacent points on the checkerboard. Besides, an iteration-based reconstruction method is proposed. 

\subsection{Calibration of Intrinsic and Extrinsic Camera Parameters}
As shown in Fig. 1, stereo vision systems consist of two cameras and two lenses. The MDM calibration method employs the most commonly used checkerboard. $m$ horizontal lines and $n$ vertical lines are arranged on the checkerboard. The two lines intersect to form a corner point. The distance between adjacent corner points is precisely known. In image acquisition, the lenses and cameras are first fixed on the optical platform, and the checkerboard is randomly placed in multiple orientations within a certain depth range. For each orientation, the two cameras capture the images of the checkerboard at the same time. A total of $2 \times w$ images of the checkerboard are acquired.

\begin{figure*}[!t]
\includegraphics[width=0.75\textwidth]{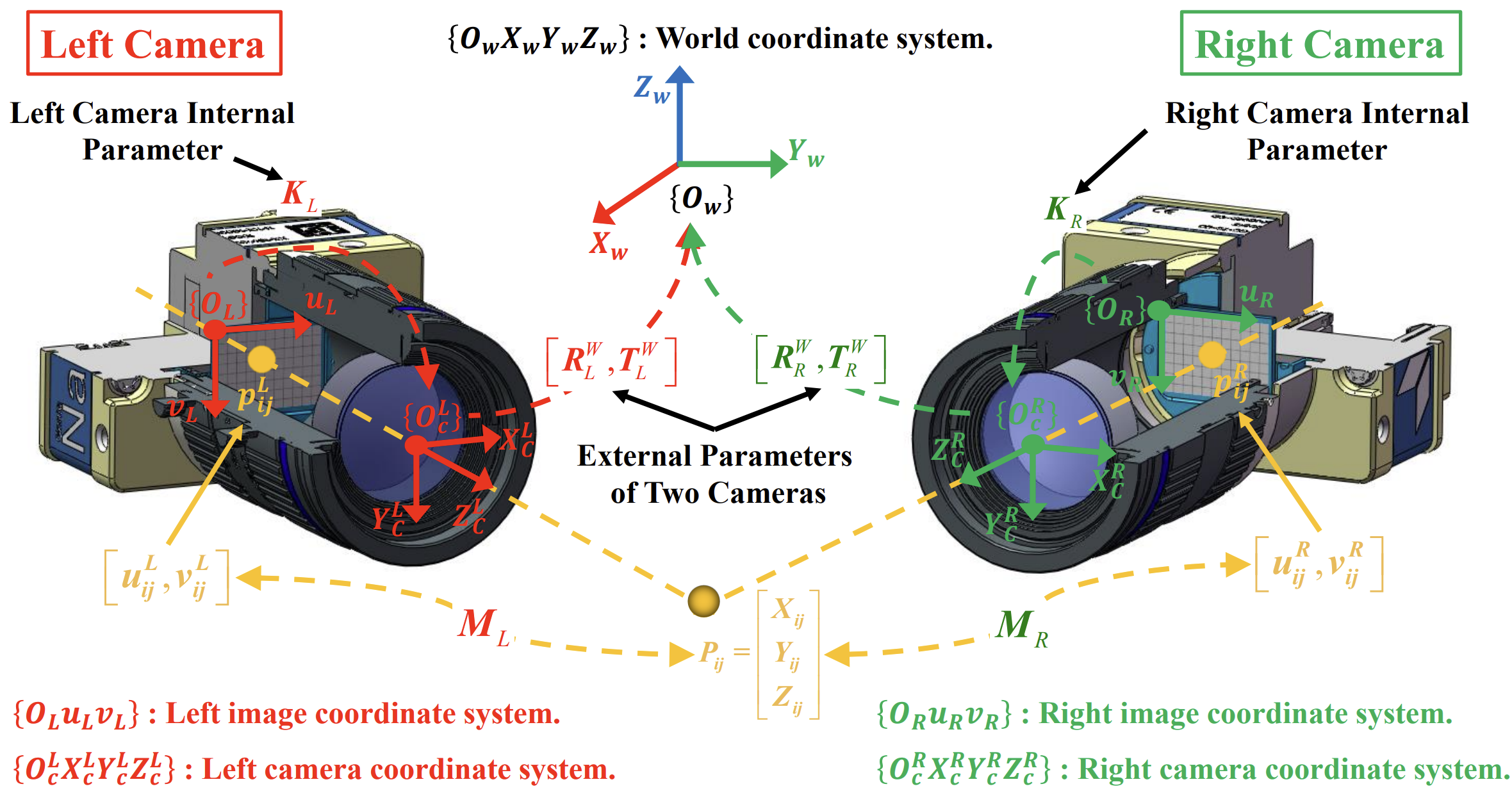}
\centering
\caption{Stereo vision system and transformations of different frames. ${O_W}$ is the origin of the world coordinate system, ${O_C^L}$ is the origin of the left camera coordinate system, ${O_C^R}$ is the origin of the right camera coordinate system, ${K_L}$ and ${K_R}$ are the intrinsic parameters of the left and right cameras, respectively. [$R_L^W$,$T_L^W$] and [$R_R^W$,$T_R^W$] are the transformation matrices between the world coordinate system and the left and right camera coordinate systems.}
\label{fig1}
\end{figure*}

For the corrected $i$-th row and $j$-th column corner point $P_{i j}$ on the checkerboard plane in the world coordinate system $O_W X_W Y_W Z_W$, the mapping relationship between its 3D reconstructed information $\left[X_{i j}, Y_{i j}, Z_{i j}\right]$ and the left image coordinate $\left(u_{i j}^L, v_{i j}^L\right)$ and that between its 3D reconstructed information $\left[X_{i j}, Y_{i j}, Z_{i j}\right]$ and the right image coordinate $\left(u_{i j}^R, v_{i j}^R\right)$ are introduced as follows:

\begin{equation}
\label{e15}
\left\{\begin{aligned}
z_L\left[\begin{array}{c}
u_{i j}^L \\
v_{i j}^L \\
1
\end{array}\right] & =\boldsymbol{K}_L\left[\boldsymbol{R}_L^W, \boldsymbol{T}_L^W\right]\left[\begin{array}{c}
X_{i j} \\
Y_{i j} \\
Z_{i j} \\
1
\end{array}\right] \\
& =\left[\begin{array}{llll}
m_{11}^L & m_{12}^L & m_{13}^L & m_{14}^L \\
m_{21}^L & m_{22}^L & m_{23}^L & m_{24}^L \\
m_{31}^L & m_{32}^L & m_{33}^L & m_{34}^L
\end{array}\right]\!\!\!\!\left[\!\!\!\begin{array}{c}
X_{i j} \\
Y_{i j} \\
Z_{i j} \\
1
\end{array}\!\!\!\right] \\
z_R\left[\begin{array}{c}
u_{i j}^R \\
v_{i j}^R \\
1
\end{array}\right] & =\boldsymbol{K}_R\left[\begin{array}{ll}
\boldsymbol{R}_R^W, \boldsymbol{T}_R^W
\end{array}\right]\left[\begin{array}{c}
X_{i j} \\
Y_{i j} \\
Z_{i j} \\
1
\end{array}\right] \\
& =\left[\begin{array}{llll}
m_{11}^R & m_{12}^R & m_{13}^R & m_{14}^R \\
m_{21}^R & m_{22}^R & m_{23}^R & m_{24}^R \\
m_{31}^R & m_{32}^R & m_{33}^R & m_{34}^R
\end{array}\right]\!\!\!\!\left[\!\!\!\begin{array}{c}
X_{i j} \\
Y_{i j} \\
Z_{i j} \\
1
\end{array}\!\!\!\right]
\end{aligned}\right.
\end{equation}

Here, $\boldsymbol{R}_L^W$ and $\boldsymbol{T}_L^W$ are the rotation matrix and the translation matrix, expressing the transformation between the left image coordinate system and the world coordinate system; $\boldsymbol{R}_R^W$ and $\boldsymbol{T}_R^W$ are the rotation matrix and the translation matrix, expressing the transformation between the right image coordinate system and the world coordinate system; $\boldsymbol{K}_L$ and  $\boldsymbol{K}_R$ are the intrinsic parameter matrices of the left and right cameras, respectively; $z_L$ and $z_R$ are the scaling factors; $m_{i j}^L$ and $m_{i j}^R$ are the values of the $i$-th row and $j$-th column in the left and right camera projection matrices, respectively; $\left(u_{i j}^L, v_{i j}^L\right)$ and $\left(u_{i j}^R, v_{i j}^R\right)$ are the corrected image coordinates in the $i$-th row and $j$-th column in the checkerboard images under the left and right image coordinate systems, respectively. The image coordinates of corner points in the checkerboard images are extracted, and the initial values of the intrinsic $\left(\boldsymbol{K}_L, \boldsymbol{K}_R\right)$ and extrinsic $\left(\boldsymbol{R}_L^W, \boldsymbol{T}_L^W, \boldsymbol{R}_R^W, \boldsymbol{T}_R^W\right)$ parameters of the cameras are calculated using the Zhang's calibration method [4]. 

\subsection{Calibration of MDM Parameters}
In this section, the radial and decentering lens distortion parameters in the MDM are roughly estimated using linear constraints. For any corrected $i$-th row and $j$-th column corner point on the checkerboard plane, 
its left image coordinate $\left(u_{i j}^L, v_{i j}^L\right)$ and right image coordinate $\left(u_{i j}^R, v_{i j}^R\right)$ are obtained by (16), 
where $A_1^L$, $A_2^L$, $A_1^R$, $A_2^R$, $B_1^L$, $B_2^L$, $B_1^R$, and $B_2^R$ are the radial lens distortion parameters; $C_1^L$, $C_2^L$,$C_1^R$, $C_2^R$, $D_1^L$, $D_2^L$, $D_1^R$, and $D_2^R$  are the decentering lens distortion parameters; $\left(\bar{u}_{i j}^L, \bar{v}_{i j}^L\right)$ and $\left(\bar{u}_{i j}^R, \bar{v}_{i j}^R\right)$ are the uncorrected corner point image coordinates in the $i$-th row and $j$-th column in the checkerboard images under the left and right image coordinate systems, respectively. $\left(u_0^L, v_0^L\right)$ and $\left(u_0^R, v_0^R\right)$ are principal point coordinates under the left and right image coordinate systems, respectively; $r_{i j}^L=\sqrt{\left(\bar{u}_{i j}^L-u_0^L\right)^2+\left(\bar{v}_{i j}^L-v_0^L\right)^2}$ and $r_{i j}^R=\sqrt{\left(\bar{u}_{i j}^R-u_0^R\right)^2+\left(\bar{v}_{i j}^R-v_0^R\right)^2}$ are distortion radii of image points under the left and right image coordinate systems, respectively; $s_{i j}^L$ and $s_{i j}^R$ are the depth information of the corner points in the $i$-th row and $j$-th column in the checkerboard images under the left and right image coordinate systems, respectively.
\begin{equation}
\label{e16}
\left\{\!\!\!\begin{aligned}
 u_{i j}^L & \!\!=\!\!\bar{u}_{i j}^L+\delta_{u_{i j}}^L \\
 v_{i j}^L & \!\!=\!\!\bar{v}_{i j}^L+\delta_{v_{i j}}^L \\
 u_{i j}^R & \!\!=\!\!\bar{u}_{i j}^R+\delta_{u_{i j}}^R \\
 v_{i j}^R & \!\!=\!\!\bar{v}_{i j}^R+\delta_{v_{i j}}^R \\
\delta_{u_{i j}}^L & \!\!=\!\!\bar{u}_{i j}^L \!\!\cdot\!\!\left[1\!\!+\!\!\left(A_1^L\!\!-\!\!B_1^L \!\!\cdot\!\! s_{i j}^L\right) \!\!\cdot\!\! r_{i j}^{L 2}\!\!+\!\!\left(A_2^L\!\!-\!\!B_2^L \!\!\cdot\!\! s_{i j}^L\right) \!\!\cdot\!\! r_{i j}^{L 4}\!\!+\!\!...\!\right] \\
& +\!\!\left[\frac{1}{C_1^L\!\!+\!\!\frac{D_1^L}{s_{i j}^L}} \!\!\cdot\!\!\left(r_{i j}^{L 2}\!\!+\!\!2 \!\!\cdot\!\! \bar{u}_{i j}^{L 2}\right)\!\!+\!\!2 \!\!\cdot\!\! \frac{1}{C_2^L\!\!+\!\!\frac{D_2^L}{s_{i j}^L}} \!\!\cdot\!\! \bar{u}_{i j}^L \!\!\cdot\!\! \bar{v}_{i j}^L\right] \\
\delta_{v_{i j}}^L & \!\!=\!\!\bar{v}_{i j}^L \!\!\cdot\!\!\left[1\!\!+\!\!\left(A_1^L\!\!-\!\!B_1^L \!\!\cdot\!\! s_{i j}^L\right) \!\!\cdot\!\! r_{i j}^{L 2}\!\!+\!\!\left(A_2^L\!\!-\!\!B_2^L \!\!\cdot\!\! s_{i j}^L\right) \!\!\cdot\!\! r_{i j}^{L 4}\!\!+\!\!...\!\right] \\
& +\!\!\left[\frac{1}{C_2^L\!\!+\!\!\frac{D_2^L}{s_{i j}^L}} \!\!\cdot\!\!\left(r_{i j}^{L 2}\!\!+\!\!2 \!\!\cdot\!\! \bar{v}_{i j}^{L 2}\right)\!\!+\!\!2 \!\!\cdot\!\! \frac{1}{C_1^L\!\!+\!\!\frac{D_1^L}{s_{i j}^L}} \!\!\cdot\!\! \bar{u}_{i j}^L \!\!\cdot\!\! \bar{v}_{i j}^L\right] \\
\delta_{u_{i j}}^R & \!\!=\!\!\bar{u}_{i j}^R \!\!\cdot\!\!\left[1\!\!+\!\!\left(A_1^R\!\!-\!\!B_1^R \!\!\cdot\!\! s_{i j}^R\right) \!\!\cdot\!\! r_{i j}^{R 2}\!\!+\!\!\left(A_2^R\!\!-\!\!B_2^R \!\!\cdot\!\! s_{i j}^R\right) \!\!\cdot\!\! r_{i j}^{R 4}\!\!+\!\!...\!\right] \\
& +\!\!\left[\frac{1}{C_1^R\!\!+\!\!\frac{D_1^R}{s_{i j}^R}} \!\!\cdot\!\!\left(r_{i j}^{R 2}\!\!+\!\!2 \!\!\cdot\!\! \bar{u}_{i j}^{R 2}\right)\!\!+\!\!2 \!\!\cdot\!\! \frac{1}{C_2^R\!\!+\!\!\frac{D_2^R}{s_{i j}^R}} \!\!\cdot\!\! \bar{u}_{i j}^R \!\!\cdot\!\! \bar{v}_{i j}^R\right] \\
 \delta_{v_{i j}}^R & \!\!=\!\!\bar{v}_{i j}^R \!\!\cdot\!\!\left[1\!\!+\!\!\left(A_1^R\!\!-\!\!B_1^R \!\!\cdot\!\! s_{i j}^R\right) \!\!\cdot\!\! r_{i j}^{R 2}\!\!+\!\!\left(A_2^R\!\!-\!\!B_2^R \!\!\cdot\!\! s_{i j}^R\right) \!\!\cdot\!\! r_{i j}^{R 4}\!\!+\!\!...\!\right] \\
& +\!\!\left[\frac{1}{C_2^R\!\!+\!\!\frac{D_2^R}{s_{i j}^R}} \!\!\cdot\!\!\left(r_{i j}^{R 2}\!\!+\!\!2 \!\!\cdot\!\! \bar{v}_{i j}^{R 2}\right)\!\!+\!\!2 \!\!\cdot\!\! \frac{1}{C_1^R\!\!+\!\!\frac{D_1^R}{s_{i j}^R}} \!\!\cdot\!\! \bar{u}_{i j}^R \!\!\cdot\!\! \bar{v}_{i j}^R\right] 
\end{aligned}\right.
\end{equation}

As shown in Fig. 2, when the corner points on the same straight lines in the left and right checkerboard images are known, the line equations determined by the corner points under the left and right image coordinate systems are described as:

\begin{equation}
\label{e17}
\left\{\begin{array}{l}
\alpha_i^L u^L-v^L+\beta_i^L=0 \\
\alpha_i^R u^R-v^R+\beta_i^R=0
\end{array}\right.
\end{equation}
where $\alpha_i^L$ and $\beta_i^L$ are the line parameters of the $i$-th line in the left checkerboard image; $\alpha_i^R$ and $\beta_i^R$ are the line parameters of the $i$-th line in the right checkerboard image; Then, (18) is obtained by substituting  $\left(\bar{u}_{i j}^L, \bar{v}_{i j}^L\right)$ and $\left(\bar{u}_{i j}^R, \bar{v}_{i j}^R\right)$ into $\left(u^L, v^L\right)$ and $\left(u^R, v^R\right)$ in (17), respectively.

\begin{equation}
\label{e18}
\left\{\begin{array}{l}
F_L\left(\alpha_i^L, \beta_i^L, \boldsymbol{K}_1^L, \boldsymbol{K}_2^L, \boldsymbol{P}_1^L, \boldsymbol{P}_2^L\right)=0 \\
F_R\left(\alpha_i^R, \beta_i^R, \boldsymbol{K}_1^R, \boldsymbol{K}_2^R, \boldsymbol{P}_1^R, \boldsymbol{P}_2^R\right)=0
\end{array}\right.
\end{equation}

$\boldsymbol{K}_1^L=\left[A_1^L, B_1^L\right]$ and $\boldsymbol{K}_2^L=\left[A_2^L, B_2^L\right]$ are the first and second-order parameters of the radial distortion under the left image coordinate system, respectively; $\boldsymbol{P}_1^L=\left[C_1^L, D_1^L\right]$ and $\boldsymbol{P}_2^L=\left[C_2^L, D_2^L\right]$ are the first and second-order parameters of the decentering distortion under the left image coordinate system, respectively; $\boldsymbol{K}_1^R=\left[A_1^R, B_1^R\right]$ and $\boldsymbol{K}_2^R=\left[A_2^R, B_2^R\right]$ are the first and second-order parameters of the radial distortion under the right image coordinate system, respectively; $\boldsymbol{P}_1^R=\left[C_1^R, D_1^R\right]$ and $\boldsymbol{P}_2^R=\left[C_2^R, D_2^R\right]$ are the first and second-order parameters of the decentering distortion under the right image coordinate system, respectively.

If there are $i$ lines in an image (left or right image) and $j$ observation points are extracted from each line, we obtain $i \times j$ equations. In these equations, there are $2 \times i+8$  variables ($2 \times i$ line parameters and 8 distortion parameters). If $i \times j>2 \times i+8$, the optimal solution of the distortion parameters is obtained. 

\begin{figure}[!t]
\begin{center}
\includegraphics[width=0.4\textwidth]{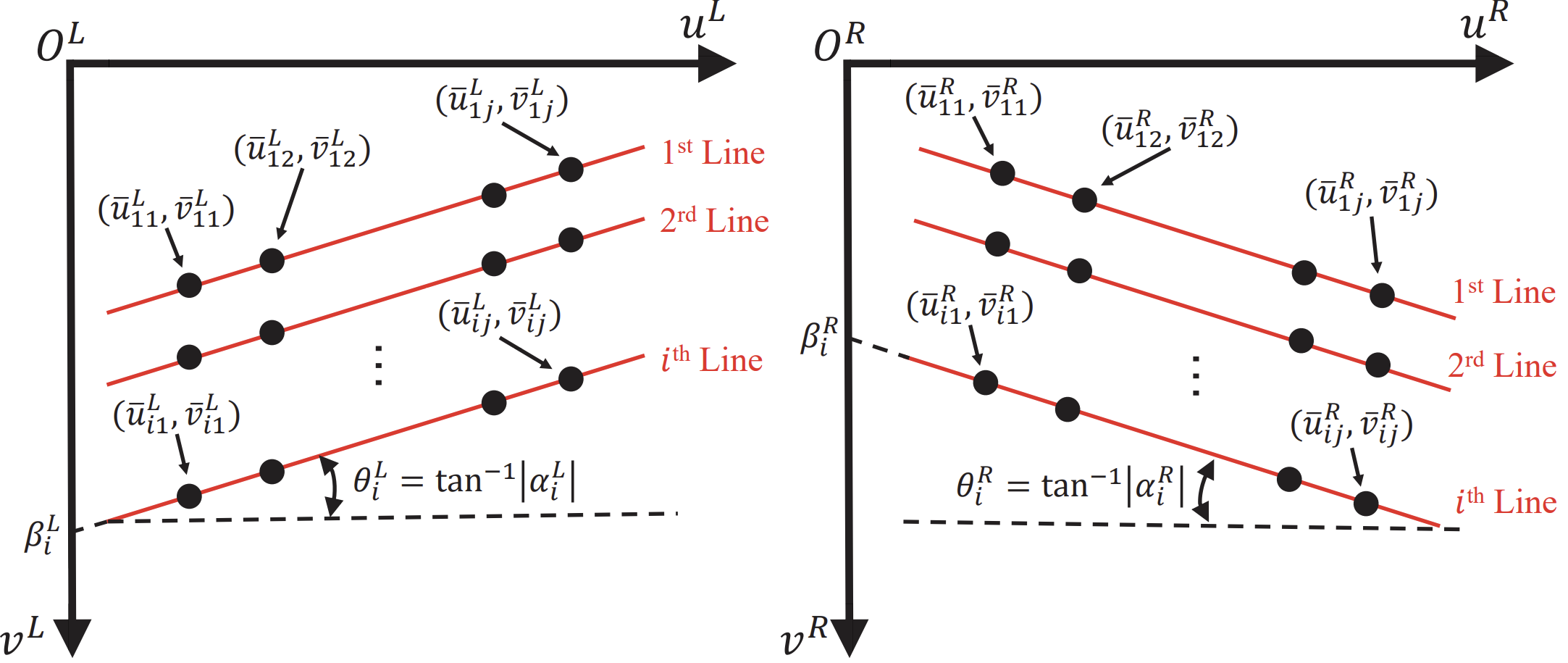}
\end{center}
\caption{Linear constraints on the left and right images}
\end{figure}

$\boldsymbol{M}_L^R=\left[\boldsymbol{R}_L^R, \boldsymbol{T}_L^R\right]$ is the 4-by-3 projection matrix, expressing the transformation relationship of the left camera coordinate system relative to the right camera coordinate system, where $\boldsymbol{R}_L^R$  and $\boldsymbol{T}_L^R$ are the rotation and translation matrices of the left camera coordinate system relative to the right camera coordinate system, respectively; $\boldsymbol{M}_R^L=\left[\boldsymbol{R}_R^L, \boldsymbol{T}_R^L\right]$ is the 4-by-3 projection matrix, expressing the transformation relationship of the right camera coordinate system relative to the left camera coordinate system, where $\boldsymbol{R}_R^L$ and $\boldsymbol{T}_R^L$ are the rotation and translation matrices of the  right camera coordinate system relative to the left camera coordinate system, respectively; $t_{i j}^L$ is the element in the $i$-th row and $j$-th column of $\boldsymbol{M}_L^R$; $t_{i j}^R$ is the element in the $i$-th row and $j$-th column of $\boldsymbol{M}_R^L$; $f_L$ and $f_R$ are the focal lengths of the left and right cameras, respectively; $s_{i j}^L=\frac{f_L \cdot\left(f_R \cdot t_{14}^L-\bar{u}_{i j}^R \cdot t_{34}^L\right)}{\bar{u}_{i j}^R \cdot\left(t_{31}^L \cdot \bar{u}_{i j}^L+t_{32}^L \cdot \bar{v}_{i j}^L+f_L \cdot t_{33}^L\right)-f_R \cdot\left(t_{11}^L \cdot \bar{u}_{i j}^L+t_{12}^L \cdot \bar{v}_{i j}^L+f_L \cdot t_{13}^L\right)}$  and $s_{i j}^R=\frac{f_R \cdot\left(f_L \cdot t_{14}^R-\bar{u}_{i j}^L \cdot t_{34}^R\right)}{\bar{u}_{i j}^L \cdot\left(t_{31}^R \cdot \bar{u}_{i j}^R+t_{32}^R \cdot \bar{v}_{i j}^R+f_R \cdot t_{33}^R\right)-f_L \cdot\left(t_{11}^R \cdot \bar{u}_{i j}^R+t_{12}^R \cdot \bar{v}_{i j}^R+f_R \cdot t_{13}^R\right)}$ are roughly estimated by substituting the calibrated intrinsic and extrinsic camera parameters, $\left(\bar{u}_{i j}^L, \bar{v}_{i j}^L\right)$ and $\left(\bar{u}_{i j}^R, \bar{v}_{i j}^R\right)$  into the equations.
\vspace{-4.5 mm}
\subsection{Overall Optimization of Intrinsic and Extrinsic Camera Parameters and MDM Parameters}
In this section, we optimize the intrinsic and extrinsic camera parameters along with the MDM parameters based on the known distances between adjacent points on the checkerboard. First, according to (15), by eliminating $z_L$ and $z_R$, we obtain (19):

\begin{equation}
\label{e19}
\begin{aligned}
& {\left[\!\!\!\begin{array}{ccc}
u_{i j}^L \!\!\cdot\!\! m_{31}^L\!\!-\!\!m_{11}^L \!\!&\!\! u_{i j}^L \!\!\cdot\!\! m_{32}^L\!\!-\!\!m_{12}^L \!\!&\!\! u_{i j}^L \!\!\cdot\!\! m_{33}^L\!\!-\!\!m_{13}^L \\
v_{i j}^L \!\!\cdot\!\! m_{31}^L\!\!-\!\!m_{21}^L \!\!&\!\! v_{i j}^L \!\!\cdot\!\! m_{32}^L\!\!-\!\!m_{22}^L \!\!&\!\! v_{i j}^L \!\!\cdot\!\! m_{33}^L\!\!-\!\!m_{23}^L \\
u_{i j}^R \!\!\cdot\!\! m_{31}^R\!\!-\!\!m_{11}^R \!\!&\!\! u_{i j}^R \!\!\cdot\!\! m_{32}^R\!\!-\!\!m_{12}^R \!\!&\!\! u_{i j}^R \!\!\cdot\!\! m_{33}^R\!\!-\!\!m_{13}^R \\
v_{i j}^R \!\!\cdot\!\! m_{31}^R\!\!-\!\!m_{21}^R \!\!&\!\! v_{i j}^R \!\!\cdot\!\! m_{32}^R\!\!-\!\!m_{22}^R \!\!&\!\! v_{i j}^R \!\!\cdot\!\! m_{33}^R\!\!-\!\!m_{23}^R
\end{array}\!\!\!\right]\!\!\!\left[\begin{array}{c}
X_{i j} \\
Y_{i j} \\
Z_{i j}
\end{array}\right]} \\
& =\left[\begin{array}{l}
m_{14}^L\!\!-\!\! u_{i j}^L \!\!\cdot\!\! m_{34}^L \\
m_{24}^L\!\!-\!\! v_{i j}^L \!\!\cdot\!\! m_{34}^L \\
m_{14}^R\!\!-\!\! u_{i j}^R \!\!\cdot\!\! m_{34}^R \\
m_{24}^R\!\!-\!\! v_{i j}^R \!\!\cdot\!\! m_{34}^R
\end{array}\right] \\
&
\end{aligned}
\end{equation}

To solve $\left[X_{i j}, Y_{i j}, Z_{i j}\right]^{T}$, (19) is rewritten as follows:

\begin{equation}
\label{e20}
\left\{\!\!\!\!\!\begin{array}{c}
\boldsymbol{A}\!=\!\left[\!\!\!\!\begin{array}{ccc}
u_{i j}^L \!\cdot\! m_{31}^L\!\!-\!\!m_{11}^L \!\!&\!\! u_{i j}^L \!\cdot\! m_{32}^L\!\!-\!\!m_{12}^L \!\!&\!\! u_{i j}^L \!\cdot\! m_{33}^L\!\!-\!\!m_{13}^L \\
v_{i j}^L \!\cdot\! m_{31}^L\!\!-\!\!m_{21}^L \!\!&\!\! v_{i j}^L \!\cdot\! m_{32}^L\!\!-\!\!m_{22}^L \!\!&\!\! v_{i j}^L \!\cdot\! m_{33}^L\!\!-\!\!m_{23}^L \\
u_{i j}^R \!\cdot\! m_{31}^R\!\!-\!\!m_{11}^R \!\!&\!\! u_{i j}^R \!\cdot\! m_{32}^R\!\!-\!\!m_{12}^R \!\!&\!\! u_{i j}^R \!\cdot\! m_{33}^R\!\!-\!\!m_{13}^R \\
v_{i j}^L \!\cdot\! m_{31}^R\!\!-\!\!m_{21}^R \!\!&\!\! v_{i j}^R \!\cdot\! m_{32}^R\!\!-\!\!m_{22}^R \!\!&\!\! v_{i j}^R \!\cdot\! m_{33}^R\!\!-\!\!m_{23}^R
\end{array}\!\!\!\!\right] \\
\boldsymbol{B}=\left[\begin{array}{l}
m_{14}^L-u_{i j}^L \cdot m_{34}^L \\
m_{24}^L-v_{i j}^L \cdot m_{34}^L \\
m_{14}^R-u_{i j}^R \cdot m_{34}^R \\
m_{24}^R-v_{i j}^R \cdot m_{34}^R
\end{array}\right] \\
\boldsymbol{H}_{i, j}^l=\left[\begin{array}{l}
X_{i j} \\
Y_{i j} \\
Z_{i j}
\end{array}\right]=\left(\boldsymbol{A}^T \boldsymbol{A}\right)^{-1} \boldsymbol{A}^T \boldsymbol{B}
\end{array}\right.
\end{equation}

Here, $\boldsymbol{A}^T$ is the transposition of matrix $\boldsymbol{A}$; $\left(\boldsymbol{A}^T \boldsymbol{A}\right)^{-1}$ is the inverse of matrix $\left(\boldsymbol{A}^T \boldsymbol{A}\right)$. Then, the objective function is described by (21):

\begin{equation}
\label{e21}
E\!\!=\!\!\min\!\!\! \sum_{l=1}^w \!\!\sum_{i=1}^{m\!-\!1} \!\!\sum_{j=1}^{n\!-\!1}\!\left(||\!| \boldsymbol{H}_{i, j}^l\!-\!\boldsymbol{H}_{i, j\!+\!1}^l\|\!\!-\!\!a|\!\!+\!\!|\| \boldsymbol{H}_{i, j}^l\!\!-\!\!\boldsymbol{H}_{i\!+\!1, j}^l \|\!-\!b \!\mid\!\right)
\end{equation}

Here, $\boldsymbol{H}_{i, j}^l$ presents the 3D information of corner points in the $i$-th row and $j$-th column in the $l$-th checkerboard image related to the intrinsic and extrinsic camera parameters and the MDM parameters; $a$ and $b$ represent the actual length and width of the rectangle on the checkerboard plane, respectively. Finally, the Levenberg-Marquardt optimization method is employed to optimize the intrinsic and extrinsic camera parameters and the MDM parameters. The optimized initial values are the intrinsic and extrinsic camera parameters calibrated by using the Zhang's calibration method and the MDM parameters calibrated using linear constraints. Based on the optimization variables, optimized initial values, and optimization objective function, the optimized values of the intrinsic and extrinsic camera parameters and the MDM parameters are obtained.

\subsection{Iteration-Based 3D Information Reconstruction Method}

For the reconstruction of the 3D information of the target, the depth information of the target in the MDM or MDM-R is unknown, which can be roughly estimated using the linear pinhole model. However, the accuracy of the estimated depth information of the target is limited, because the linear pinhole model does not consider the lens distortion. Thus, in this section, an iteration-based 3D information reconstruction method is proposed. 

\floatname{algorithm}{Algorithm 1} 
\renewcommand{\algorithmicrequire}{\textbf{Input:}} 
\renewcommand{\algorithmicensure}{\textbf{Output:}} 
\linespread{1.2}
\renewcommand{\thealgorithm}{:}
    \begin{algorithm}[H]
        \caption{Iteration-based 3D Reconstruction Method} 
        \begin{algorithmic}[1] 
            \Require Intrinsic and extrinsic stereo camera parameters, MDM/MDM-R parameters, sufficiently small $\epsilon$
            \Ensure 3D reconstructed information of the target $\left[X_{i j}^{k},Y_{i j}^{k},Z_{i j}^{k}\right]^{\text{T}}$

            \State $k\Leftarrow1$
            
            \State $\!\boldsymbol{S}\!\!\Leftarrow \!\!\left(\!\!\boldsymbol{K}\!\!_L\!,\! \boldsymbol{K}\!\!_R\!,\!\boldsymbol{R}\!_L^W\!\!,\! \boldsymbol{T}\!_L^W\!\!,\! \boldsymbol{R}\!_R^W\!,\! \boldsymbol{T}\!_R^W\!,\! \boldsymbol{K}\!_1^L\!,\! \boldsymbol{K}\!_2^L\!,\! \boldsymbol{P}\!_1^L\!,\! \boldsymbol{P}\!_2^L\!,\! \boldsymbol{K}\!_1^R\!,\! \boldsymbol{K}\!_2^R\!,\! \boldsymbol{P}\!_1^R\!,\! \boldsymbol{P}\!_2^R\!\!\right)$

            \State $^{k}\!s_{ij}^{L} \!\!\Leftarrow\!\!\frac{f_L \cdot\left(f_R \cdot t_{14}^L-\bar{u}_{i j}^R \cdot t_{34}^L\right)}{\bar{u}_{i j}^R \cdot\left(t_{31}^L \cdot \bar{u}_{i j}^L+t_{32}^L \cdot \bar{v}_{i j}^L+f_L \cdot t_{33}^L\right)-f_R \cdot\left(t_{11}^L \cdot \bar{u}_{i j}^L+t_{12}^L \cdot \bar{v}_{i j}^L+f_L \cdot t_{13}^L\right)}$
            \State $^{k}\!s_{ij}^{R} \!\!\Leftarrow\!\!\frac{f_R \cdot\left(f_L \cdot t_{14}^R-\bar{u}_{i j}^L \cdot t_{34}^R\right)}{\bar{u}_{i j}^L \cdot\left(t_{31}^R \cdot \bar{u}_{i j}^R+t_{32}^R \cdot \bar{v}_{i j}^R+f_R \cdot t_{33}^R\right)-f_L \cdot\left(t_{11}^R \cdot \bar{u}_{i j}^R+t_{12}^R \cdot \bar{v}_{i j}^R+f_R \cdot t_{13}^R\right)}$
            
            \While{$\delta \geq \epsilon$}

                    \State $[X_{i j}^{k},Y_{i j}^{k},Z_{i j}^{k}]^{T}\Leftarrow \boldsymbol{H}_{i, j}^l(\boldsymbol{S},^{k}\!s_{ij}^{L}, ^{k}\!s_{ij}^{R})$
                    \State $^{k+1}\!s_{ij}^{L}\Leftarrow \left[ 0,0,1 \right]\left[ \boldsymbol{R}_L^W,\boldsymbol{T}_L^W \right]\left[X_{i j}^{k},Y_{i j}^{k},Z_{i j}^{k},1\right]^{T} $
                    \State $^{k+1}\!s_{ij}^{R}\Leftarrow \left[ 0,0,1 \right]\left[ \boldsymbol{R}_R^W,\boldsymbol{T}_R^W \right]\left[X_{i j}^{k},Y_{i j}^{k},Z_{i j}^{k},1\right]^{T}$
                    \State $[X_{i j}^{k+1},Y_{i j}^{k+1},Z_{i j}^{k+1}]^{T}\Leftarrow \boldsymbol{H}_{i, j}^l(\boldsymbol{S},^{k+1}\!s_{ij}^{L}, ^{k+1}\!s_{ij}^{R})$
                     \State $\delta \Leftarrow \left\| \left[X_{i j}^{k+1},Y_{i j}^{k+1},Z_{i j}^{k+1}\right]^{T}-\left[X_{i j}^{k},Y_{i j}^{k},Z_{i j}^{k}\right]^{T} \right\|$
                    \State $k\Leftarrow k+1$
            \EndWhile   

            \State \Return $\left[X_{i j}^{k},Y_{i j}^{k},Z_{i j}^{k}\right]^{\text{T}}$
        \end{algorithmic}
    \end{algorithm}

Using the proposed iteration-based 3D information reconstruction method, the depth information of the target in the MDM or MDM-R is updated iteratively, which improves the overall reconstruction accuracy. The detailed implementation is shown in Algorithm 1. In the above iteration-based 3D information reconstruction method, the inputs are the intrinsic and extrinsic stereo camera parameters, the MDM/MDM-R parameters, and a sufficiently small $\varepsilon$. First, the initial depth information of the target is estimated using the pinhole model; consequently, the iteration algorithm runs, and the depth information of the target updates accordingly. When $\delta<\varepsilon$, the process of iteration is finished, and more accurate 3D information of the target is reconstructed. 

\section{Experiments and Results}

\subsection{Experimental Setup}
As shown in Fig. 3, the setup of the stereo vision system comprised two cameras (Nano-M2420, DALSA, Canada), two lenses (GX-5M-1216, GXTECH, China), and a computer. The resolution of the camera was $\text{2464} \times \text{2056}$ pixels, and the pixel size was $\text{3.45} \mu \mathrm{m} \times \text{3.45} \mu \mathrm{m}$. The focal length of each lens was 12 mm. The images were transmitted from the camera to the computer using the TCP/IP protocol, and the frequency of each camera was 20 Hz. The cameras were mounted on a beam-like holder with a quick-release buckle, and the holder was mounted on the optical platform.

\begin{figure}[!t]
\centering
\includegraphics[width=0.5\textwidth]{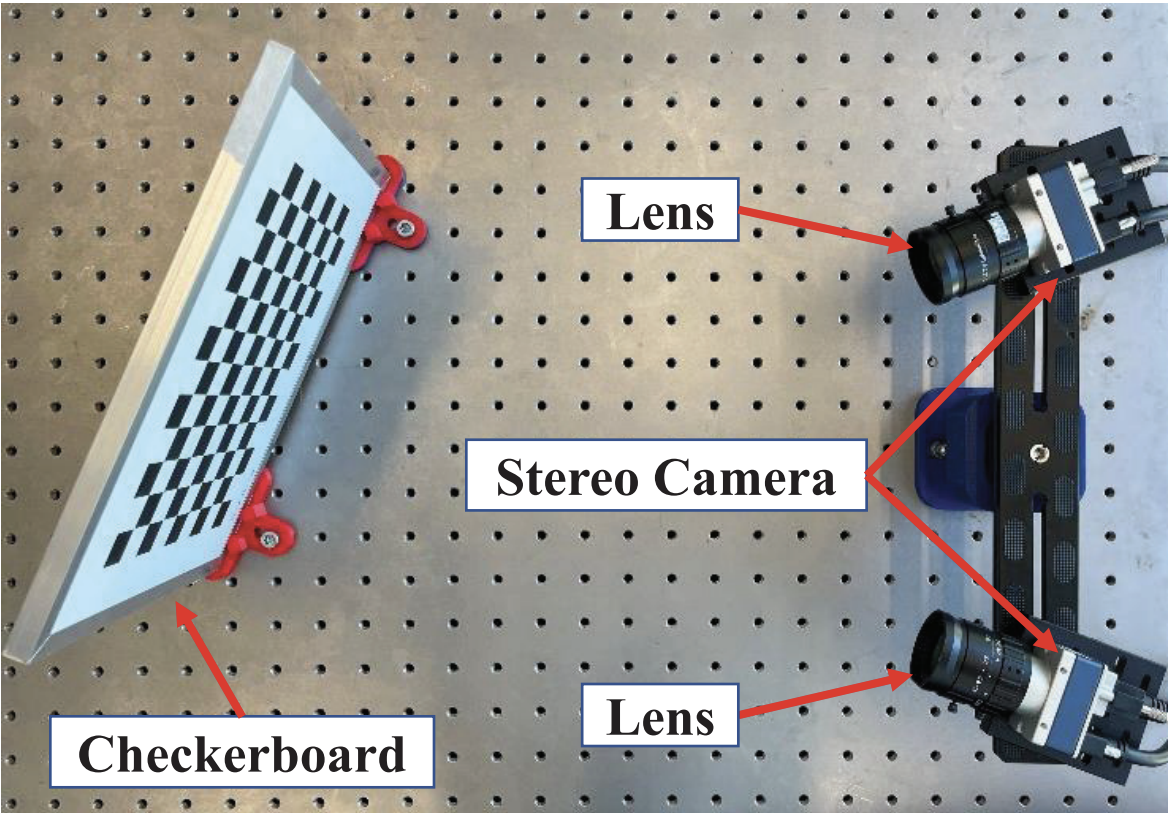}
\captionsetup{name={Fig.}, labelsep=period, singlelinecheck=off} 
\caption{Experimental setup}
\label{fig3}
\end{figure}

As shown in Fig. 4, the checkerboard plane contains 12×9 (150×105 mm) squares. There are 8 horizontal and 11 vertical lines distributed on the checkerboard, and the lines intersect to form a total of 88 corner points. The distances between adjacent corner points are 15 mm with a positioning accuracy of 0.005 mm.

\begin{figure}[!t]
\centering
\includegraphics[width=0.5\textwidth]{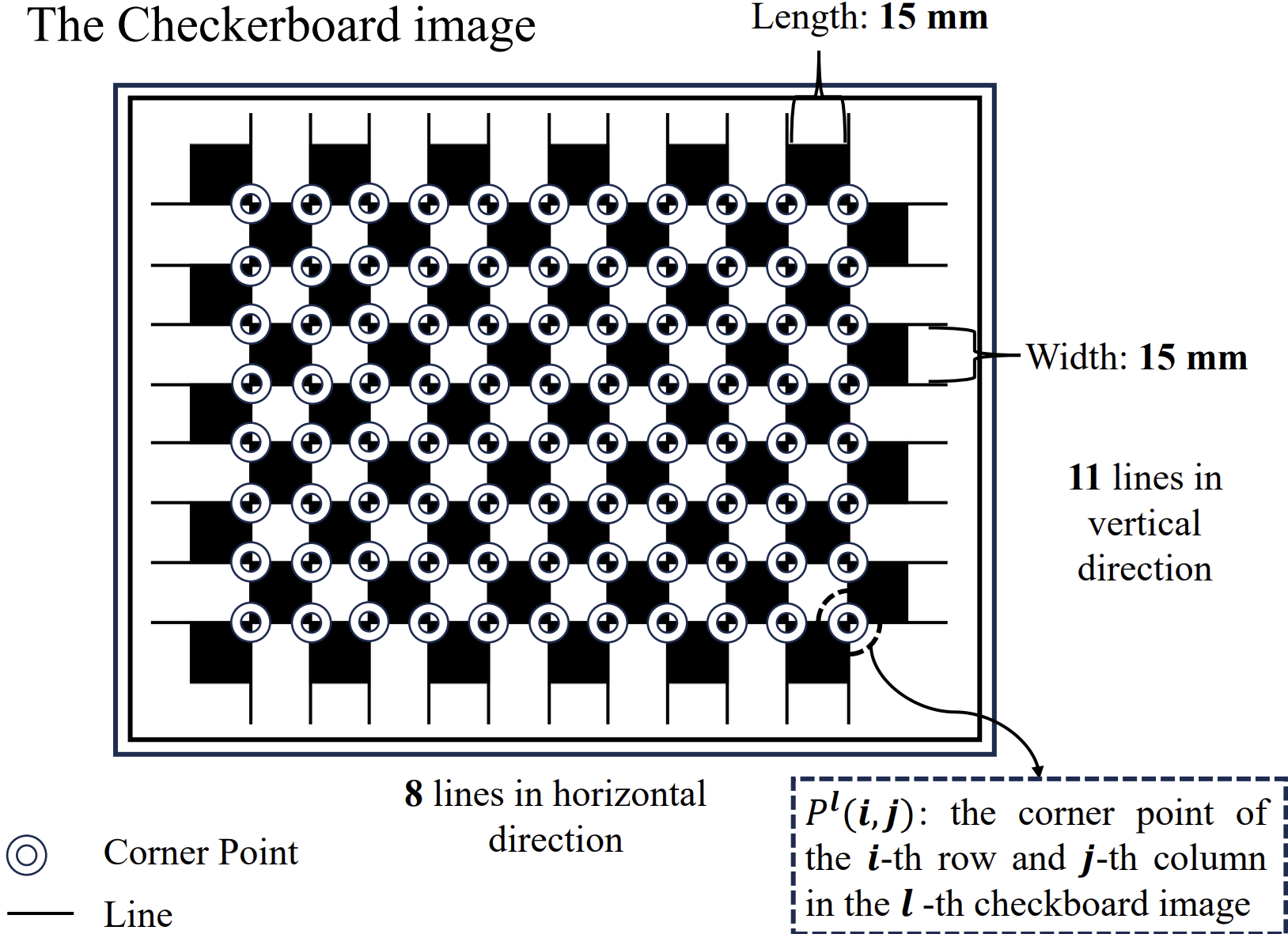}
\captionsetup{name={Fig.}, labelsep=period, singlelinecheck=off} 
\caption{Schematic diagram of the checkerboard layout}
\end{figure}

\begin{figure}[!t]
\centering  
\subfigure[]{   
\begin{minipage}{9cm}
    
\includegraphics[width=0.9\textwidth]{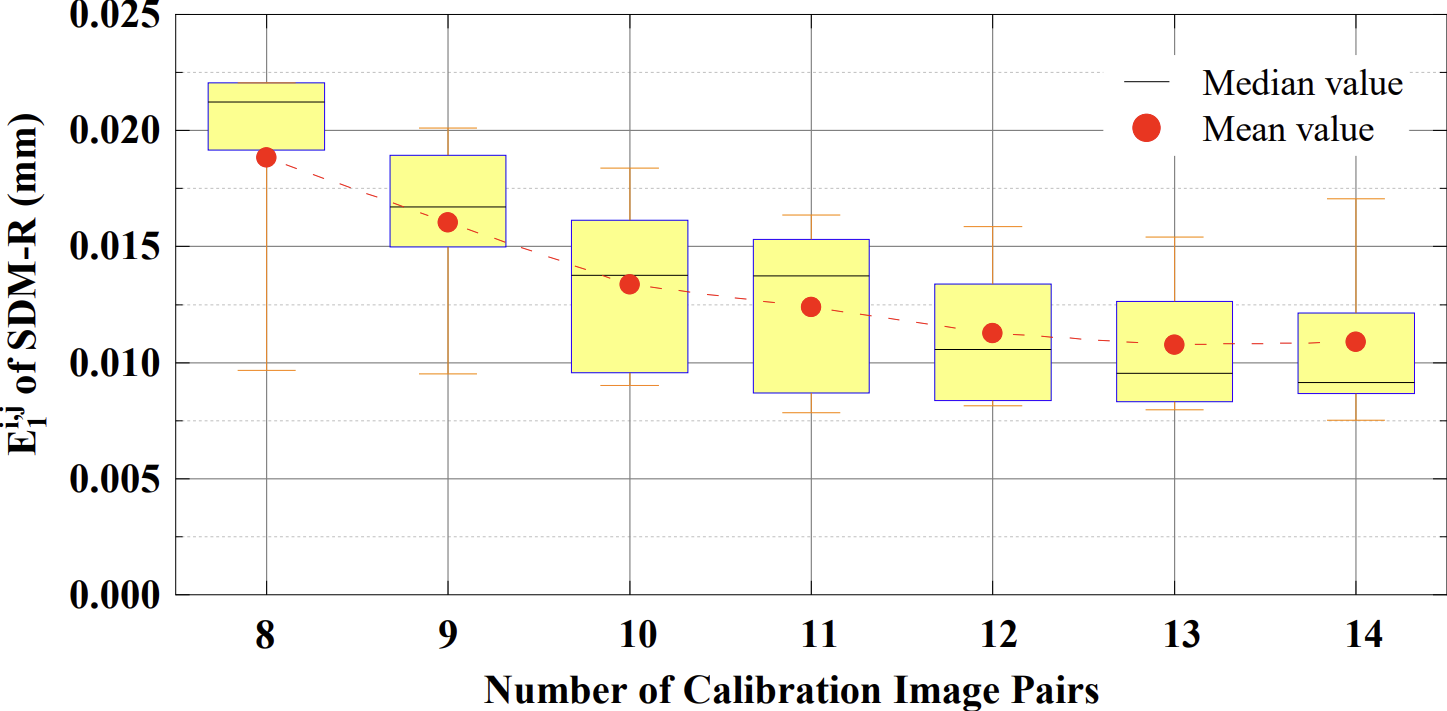}  
\end{minipage}
}
\subfigure[]{ 
\begin{minipage}{9cm}
    
\includegraphics[width=0.9\textwidth]{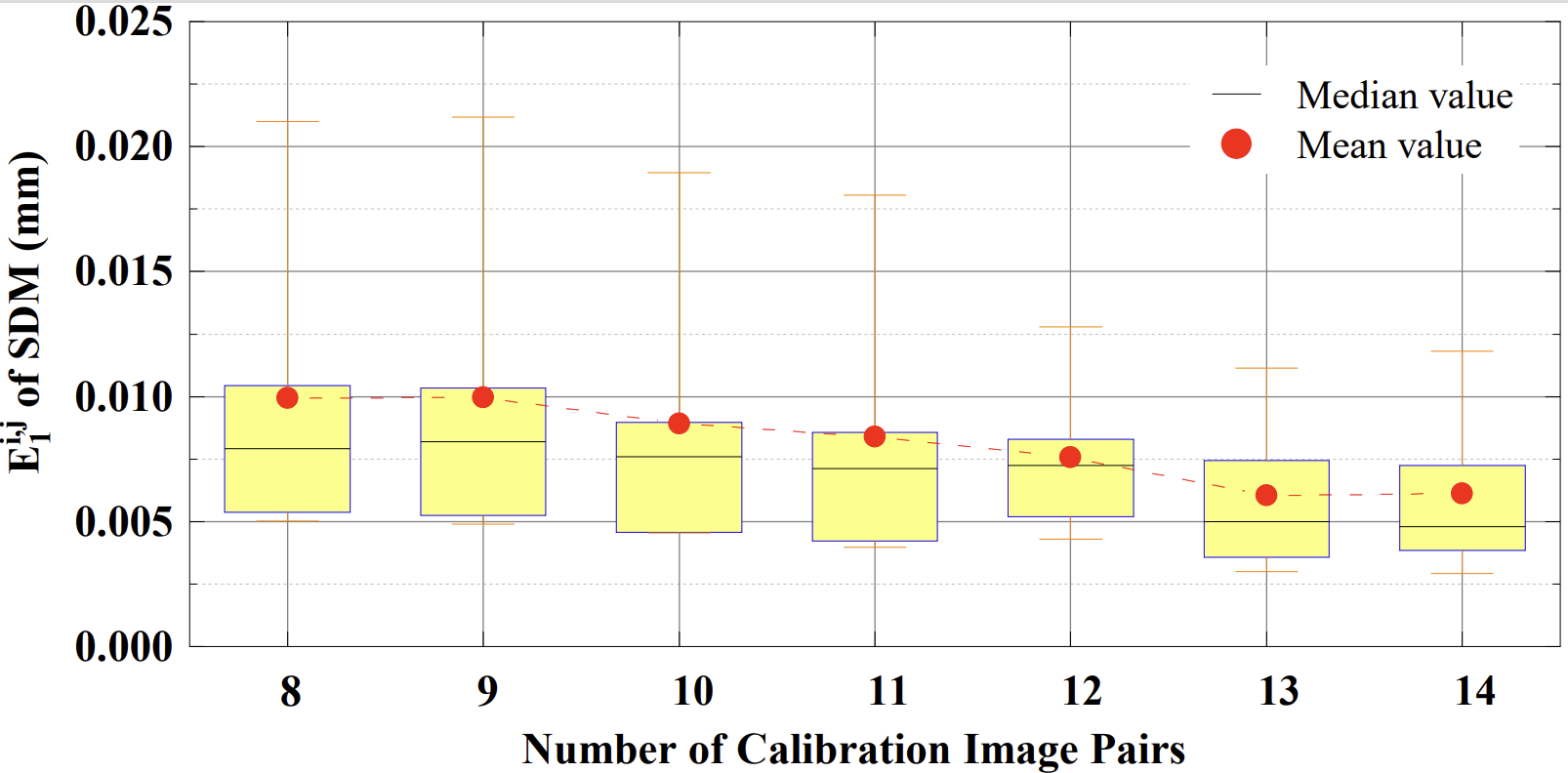}
\end{minipage}
}
\caption{Relationship between the calibration accuracy and the number of calibration image pairs. (a) Relationship between the number of calibration images and $E_1^{i, j}$ of the MDM-R. (b) Relationship between the number of calibration images and  $E_1^{i, j}$ of the MDM.}    
\label{fig:1}    
\end{figure}

\subsection{Relationship Between the Number of Calibration Image Pairs and the Calibration Accuracy of the MDM}
We conducted several experiments to demonstrate the relationship between the number of calibration image pairs and the calibration accuracy of the MDM. The stereo vision system was used to acquire 8, 9, 10, 11, 12, 13, and 14 pairs of calibration images with different checkerboard orientations, respectively. Then, the proposed method was used to calibrate the MDM-R and the MDM using different pairs of calibration images. Finally, another 5 pairs of images with different checkerboard orientations were acquired to validate the accuracy of the MDM-R and the MDM, respectively. 

The calibration error was defined as:

\begin{equation}
\label{e22}
E_1^{i, j}=\left|\|\boldsymbol{H}_{i, j}^l-\boldsymbol{H}_{i, j+1}^l \|-a\right|
\end{equation}
where $i=1,2, \ldots, 8$ ; $j=1,2, \ldots, 10$; $l$ represents different calibration images. Fig. 5(a) shows the relationship between the number of calibration image pairs and $E_1^{i, j}$  for a total of 5 tested calibration image pairs of the MDM-R. From Fig. 5(a), the $E_1^{i, j}$ of the MDM-R tended to decrease as the number of calibration image pairs increased. In particular, the $E_1^{i, j}$ decreased rapidly when the number of calibration image pairs ranged from 8 to 12. However, it decreased slowly when the number of calibration image pairs exceeded 13. Fig. 5(b) shows the relationship between the number of calibration image pairs and the $E_1^{i, j}$ of the MDM. The results indicated that $E_1^{i, j}$ decreased slowly when the number of calibration image pairs ranged from 8 to 14, but the maximum error decreased rapidly. Based on the results shown in Fig. 5, the calibration accuracy was satisfied for the MDM-R and the MDM with 13 pairs of calibration images. 

\begin{table*}[]
\centering
\caption{calibration errors and calibration time with different lens distortion models.}
\begin{tabular}{cccccc|ccc}
\hline
\multicolumn{6}{c}{Calibration Errors with Different Distortion Models ($\mu$m)}                                                                                                                                                                                                                                                                                       & \multicolumn{3}{c}{Calibration Time (s)}                                           \\ \hline
\multirow{2}{*}{\begin{tabular}[c]{@{}c@{}}Brown's radial \\ distortion model\end{tabular}} & \multirow{2}{*}{\begin{tabular}[c]{@{}c@{}}Li's radial \\ distortion model\end{tabular}} & \multicolumn{1}{c|}{\multirow{2}{*}{MDM-R}} & \multirow{2}{*}{Brown's model} & \multirow{2}{*}{Li's model} & \multirow{2}{*}{MDM} & \multirow{2}{*}{Brown's model} & \multirow{2}{*}{Li's model} & \multirow{2}{*}{MDM} \\
                                                                                               &                                                                                             & \multicolumn{1}{c|}{}                       &                                &                             &                      &                                &                             &                      \\ \hline
167.5                                                                                          & 79.0                                                                                        & \multicolumn{1}{c|}{35.3}                   & 73.1                           & 43.5                        & 18.9                 & 342                            & 1328                        & 351                  \\ \hline
\end{tabular}
\label{Table 3}
\end{table*}

\subsection{Calibration Accuracy Comparison}


In this section, the calibration accuracy of the MDM-R and MDM was verified. Besides, the efficiency of the proposed calibration method was confirmed.

In the experiment, firstly, the checkerboard was placed within a depth range of 500 mm to 900 mm from the cameras. Secondly, according to Section II.B, 13 pairs of calibration images with different checkerboard orientations were acquired. Thirdly, our proposed calibration method was used to calibrate the MDM-R and MDM, while Zhang's calibration method was adopted for calibrating Brown's radial distortion model and Brown's distortion model. In addition, according to Li's calibration method, Li's radial distortion model and Li's distortion model were calibrated. During the calibration of these lens distortion models, the calibration time was recorded simultaneously. The calibration errors and corresponding calibration time with different distortion models are shown in Table II.

From the results in Table II, the calibration error of the MDM-R decreased by 55.32\% and 78.93\% compared to that of Li's radial distortion model and Brown's radial distortion model, respectively. Additionally, the calibration error of the MDM decreased by 56.55\% and 74.15\% compared to that of Li's distortion model and Brown's distortion model, respectively.
Without a specific requirement of the checkerboard's orientation compared with Li's calibration method, the calibration time of the MDM and Brown's distortion model was roughly four times shorter than that of Li's distortion model. This demonstrated that our proposed calibration method had higher calibration efficiency. Moreover, depth information was considered in the MDM and Li's distortion model, resulting in higher calibration accuracy compared to Brown's distortion model. Due to more parameters needing to be calibrated in Li's distortion model compared with the MDM, the calibration accuracy of Li's distortion model was lower than that of the MDM.

\begin{figure}[!t]
\begin{center}
\includegraphics[width=0.48\textwidth]{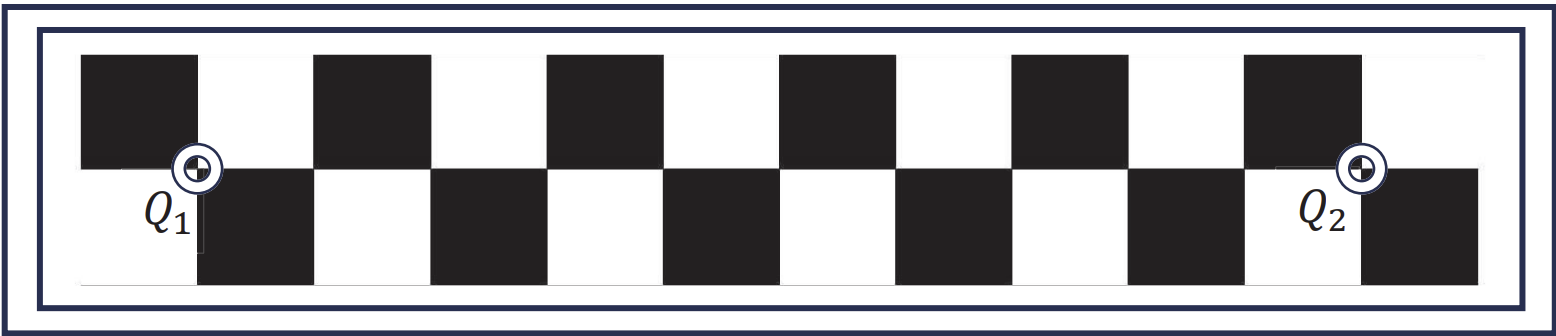}
\end{center}
\caption{The one-dimensional (1D) target}
\end{figure}
\vspace{-5 mm}
\subsection{Measurement Accuracy Comparison}
In this section, the measurement accuracy of the MDM and MDM-R was verified using a 1D target, and the proposed iteration-based reconstruction method was also validated. As shown in Fig. 6, the distance between the corner points $Q_1$ and $Q_2$ on the 1D target was precisely known as 150 mm. The positioning accuracy of the 1D target was 0.005 mm. 

In the experiment, the 1D target was placed within the depth range of 680 mm to 800 mm sequentially, and the image pairs of the 1D target were acquired accordingly for measurement accuracy verification. Then, the distance between the corner points $Q_1$ and $Q_2$ was measured at different depths using the stereo vision system with different lens distortion models and 3D reconstruction methods. 

Brown's radial distortion  model and Li's radial distortion model were calibrated using Zhang's calibration method and Li's calibration method, respectively. Brown's and Li's distortion models were calibrated using Zhang's calibration method and Li's calibration method, respectively. The MDM-R and MDM were calibrated by the proposed calibration method.
The non-iteration reconstruction method was employed for Brown's radial distortion model, as well as Brown's distortion model. For conducting 3D reconstruction, the depth information in Li's radial distortion model and Li's distortion model was roughly estimated. However, the iteration-based reconstruction method was adopted to address the same problem in MDM-R and MDM. The initial depth information was obtained by pin-hole model, and Algorithm 1 in Section III.B was used to iteratively calculate more accurate depth information. The 3D reconstruction error using the iteration-based reconstruction method was compared with that using the non-iteration reconstruction method.
The measurement errors with these lens distortion models and 3D reconstruction methods at different depths are shown in Table III.

From the results in Table III, it was obvious that the measurement errors using Li's radial distortion model and Li's distortion model were larger than those using the MDM-R and MDM. Besides, the measurement errors using Brown's radial distortion model and Brown's distortion model were larger than those using Li's radial distortion model and Li's distortion model. Due to more accurate depth information calculated with each iteration, the measurement accuracy by using the iteration-based reconstruction method was higher.
Compared with the non-iteration reconstruction method for the MDM, the measurement errors using the iteration-based reconstruction method were decreased by 9.08\% (on average). The results in Table III not only verified the measurement accuracy of our proposed distortion model calibrated by the proposed calibration method but also demonstrated the validity of the iteration-based reconstruction method.

\begin{table*}[]
\centering
\caption{Measurement errors with different distortion models at different depths.}
\begin{tabular}{ccccccccc}
\hline
\multirow{4}{*}{\begin{tabular}[c]{@{}c@{}}$\left\|Q_1-Q_2\right\|$\\ with different\\ depths (mm)\end{tabular}} & \multicolumn{8}{c}{Measurement Errors with Different Distortion Models ($\mu$m)}                                                                                                                                                                                                                                                                                                                                                                                                                                                                                                                                                                                                                                                                     \\ \cline{2-9} 
                                                                                              & \multirow{3}{*}{\begin{tabular}[c]{@{}c@{}}Brown's radial\\ distortion model\end{tabular}} & \multirow{3}{*}{\begin{tabular}[c]{@{}c@{}}Li's radial\\ distortion model\end{tabular}} & \multicolumn{2}{c}{MDM-R}                                                                                                                                                           & \multirow{3}{*}{\begin{tabular}[c]{@{}c@{}}Brown's \\ model\end{tabular}} & \multirow{3}{*}{\begin{tabular}[c]{@{}c@{}}Li's\\ model\end{tabular}} & \multicolumn{2}{c}{MDM}                                                                                                                                                             \\ \cline{4-5} \cline{8-9} 
                                                                                              &                                                                                                   &                                                                                                & \multirow{2}{*}{\begin{tabular}[c]{@{}c@{}}Non-iteration\\ reconstruction\end{tabular}} & \multirow{2}{*}{\begin{tabular}[c]{@{}c@{}}Iteration-based\\ reconstruction\end{tabular}} &                                                                           &                                                                       & \multirow{2}{*}{\begin{tabular}[c]{@{}c@{}}Non-iteration\\ reconstruction\end{tabular}} & \multirow{2}{*}{\begin{tabular}[c]{@{}c@{}}Iteration-based\\ reconstruction\end{tabular}} \\
                                                                                              &                                                                                                   &                                                                                                &                                                                                         &                                                                                           &                                                                           &                                                                       &                                                                                         &                                                                                           \\ \hline
\multicolumn{1}{c|}{680}                                                                      & 384.2                                                                                             & 97.1                                                                                           & 41.2                                                                                    & \multicolumn{1}{c|}{39.1}                                                                 & 96.0                                                                      & 56.9                                                                  & 24.8                                                                                    & 22.9                                                                                      \\
\multicolumn{1}{c|}{700}                                                                      & 397.9                                                                                             & 99.5                                                                                           & 46.6                                                                                    & \multicolumn{1}{c|}{44.1}                                                                 & 99.5                                                                      & 58.4                                                                  & 26.6                                                                                    & 24.1                                                                                      \\
\multicolumn{1}{c|}{720}                                                                      & 414.1                                                                                             & 102.3                                                                                          & 51.1                                                                                    & \multicolumn{1}{c|}{48.2}                                                                 & 103.0                                                                     & 60.6                                                                  & 27.5                                                                                    & 25.3                                                                                      \\
\multicolumn{1}{c|}{740}                                                                      & 422.8                                                                                             & 105.6                                                                                          & 56.7                                                                                    & \multicolumn{1}{c|}{52.9}                                                                 & 105.6                                                                     & 61.9                                                                  & 29.3                                                                                    & 26.4                                                                                      \\
\multicolumn{1}{c|}{760}                                                                      & 459.1                                                                                             & 110.2                                                                                          & 61.0                                                                                    & \multicolumn{1}{c|}{58.7}                                                                 & 110.0                                                                     & 64.5                                                                  & 32.5                                                                                    & 29.6                                                                                      \\
\multicolumn{1}{c|}{780}                                                                      & 474.4                                                                                             & 115.1                                                                                          & 67.2                                                                                    & \multicolumn{1}{c|}{63.6}                                                                 & 113.8                                                                     & 67.3                                                                  & 34.8                                                                                    & 31.9                                                                                      \\
\multicolumn{1}{c|}{800}                                                                      & 497.2                                                                                             & 119.6                                                                                          & 71.4                                                                                    & \multicolumn{1}{c|}{67.3}                                                                 & 115.4                                                                     & 69.9                                                                  & 38.2                                                                                    & 34.1                                                                                      \\ \hline
\end{tabular}
\end{table*}

\begin{table*}
\centering
\caption{Calibration results of the kinematic parameters of the robotic arm.}
\scriptsize
\setlength{\tabcolsep}{0.8mm}{
\begin{tabular}{c|cccc|cccc|cccc|cccc}
\hline \multirow{3}{*}{ Link } & \multicolumn{4}{c|}{$a_i(\mathrm{~mm})$} & \multicolumn{4}{c|}{$\alpha_i\left({ }^{\circ}\right)$} & \multicolumn{4}{c|}{$d_i(\mathrm{~mm})$} & \multicolumn{4}{c}{$\theta_i\left({ }^{\circ}\right)$}  \\
 & \begin{tabular}{c} 
Zhang's \\
method
\end{tabular} & \begin{tabular}{c} 
MDM 
\end{tabular} & \begin{tabular}{c} 
Li's \\
method
\end{tabular} & \begin{tabular}{c} 
Nominal \\
value
\end{tabular} & \begin{tabular}{c} 
Zhang's \\
method
\end{tabular} & \begin{tabular}{c} 
MDM
\end{tabular} & \begin{tabular}{c} 
Li's \\
method
\end{tabular} & \begin{tabular}{c} 
Nominal \\
value
\end{tabular} & \begin{tabular}{c} 
Zhang's \\
method
\end{tabular} & \begin{tabular}{c} 
MDM
\end{tabular} & \begin{tabular}{c} 
Li's \\
method
\end{tabular} & \begin{tabular}{c} 
Nominal \\
value
\end{tabular} & \begin{tabular}{c} 
Zhang's \\
method
\end{tabular} & \begin{tabular}{c} 
MDM
\end{tabular} & \begin{tabular}{c} 
Li's \\
method
\end{tabular} & \begin{tabular}{c} 
Nominal \\
value

\end{tabular} \\
\hline 1 & 0.80	& 0.35 & 0.81 & 0.00	& 0.87 & 1.42 &	0.24 & 0.00 &	365.88 &	366.29 &	365.19 & 365.00 &	0.00 &	0.00 &	0.00 & 0.00\\
 2 & 1.06 & 1.82 & 0.46	& 0.00 & 91.07	& 91.53	& 90.31	& 90.00 & 65.51	& 66.39	& 65.18	& 65.00 & 0.00	& 0.00	& 0.00 & 0.00 \\
 3 & 0.84	& 0.54	& 0.67	& 0.00 & -91.27 & -90.54 & -90.56 & -90.00 & 396.49	& 395.19 & 396.13 & 395.00 & 0.00 & 0.00	& 0.00 & 0.00\\
 4 & 21.46	& 20.38	& 20.52	& 220.00 & 91.03	& 90.59	& 90.16	& 90.00 & -53.98 & -53.52 & -54.24 & -55.00 & 0.00 & 0.00 & 0.00  & 0.00\\
 5 & -18.94	& -19.13 & -19.90 & -20.00 & -90.24 & -91.92	& -91.17 & -90.00 & 385.27 & 386.95 & 385.71	& 385.00 & 0.00 & 0.00 & 0.00 & 0.00\\
 6 & 1.45	& 1.82	& 0.31	& 0.00 & 90.21	& 90.43	& 91.23	& 90.00 & 100.35 & 101.51 & 100.19	& 100.00 & 91.46	& 90.73	& 91.37 & 90.00\\
 7 & 111.35	& 110.38	& 110.16	& 110.00 & 90.52	& 90.76	& 90.64	& 90.00 & 135.10 & 134.41 & 135.11 & 135.00 & 0.00 & 0.00 & 0.00 & 0.00\\
\hline
\end{tabular}
}
\end{table*}

\subsection{Robotic Arm Kinematic Parameters Calibration}

In this section, we applied our proposed stereo vision system to the kinematic parameters calibration of a robotic arm. 
As shown in Fig. 7, the experimental setup comprised a seven-degree-of-freedom robotic arm (Flexiv Robotics, Rizon 4), a stereo vision system, and a spherical marker. The stereo vision system consisted of two cameras, two lenses, and two LED lights. The marker had a diameter of 20 mm and was attached to the end-effector of the robotic arm.
\begin{figure}[!t]
    \centering
    \includegraphics[width=0.37\textwidth]{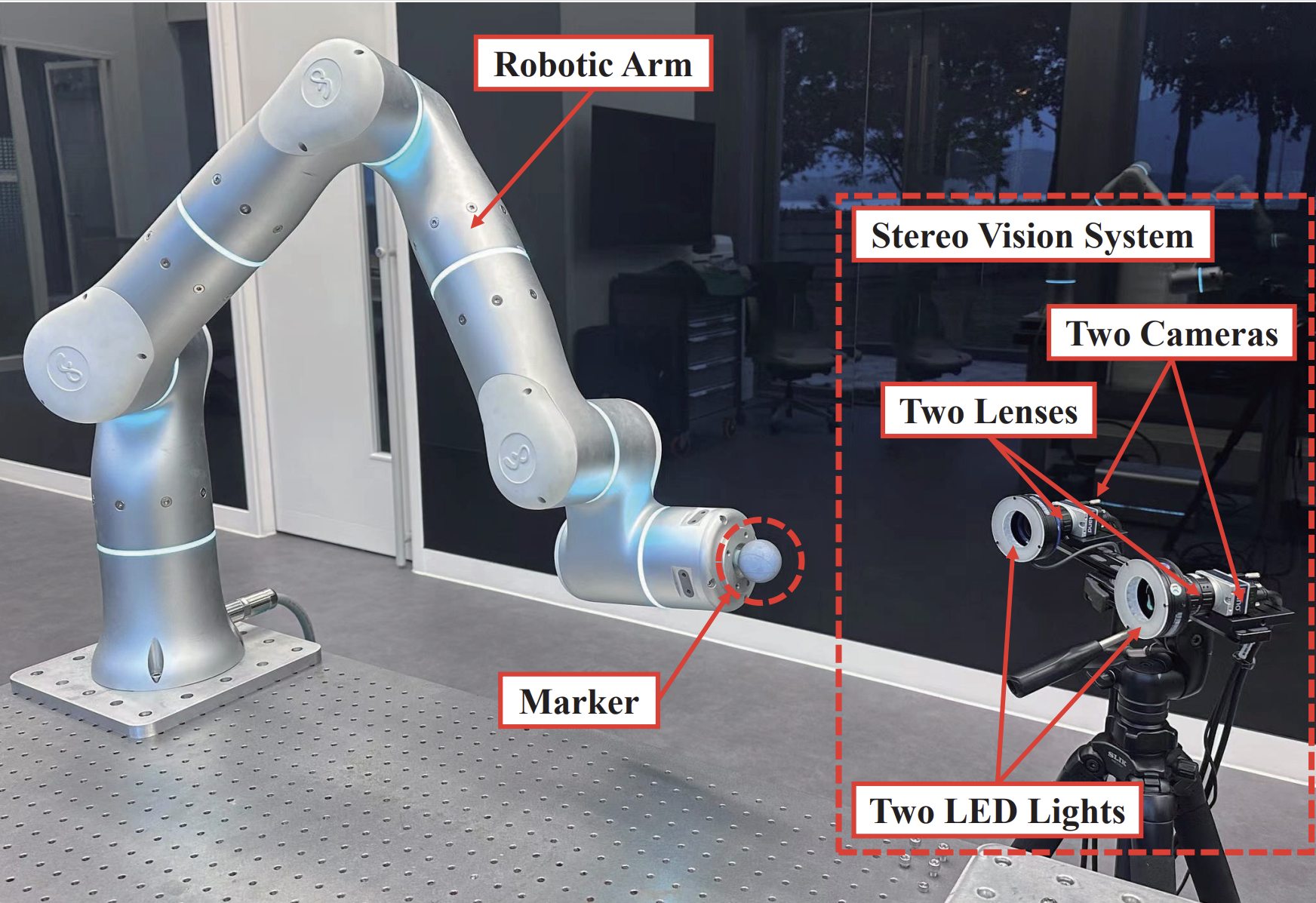}
    \caption{Experimental setup for kinematic parameters calibration of the robotic arm}
    \label{fig:exp_setup}
\end{figure}
\par
In the experiment, firstly, Brown's distortion model, Li's distortion model, and MDM were used to calibrate the stereo vision system. From the results in Table III, the measurement accuracy of the stereo vision system calibrated by MDM was higher than calibrated by Brown's distortion model and Li's distortion model. Secondly, the stereo vision system with different distortion models was used to calibrate the kinematic parameters of the robotic arm. The method in [20] was employed to calibrate the kinematic parameters of the robotic arm. Thirdly, to validate the performance of the kinematic parameters calibration, the end-effector of the robotic arm was moved to the 50 desired positions. The position of the end-effector of the robotic arm was measured by the stereo vision system.
\begin{figure}[H]
    \centering
    \includegraphics[width=0.4\textwidth]{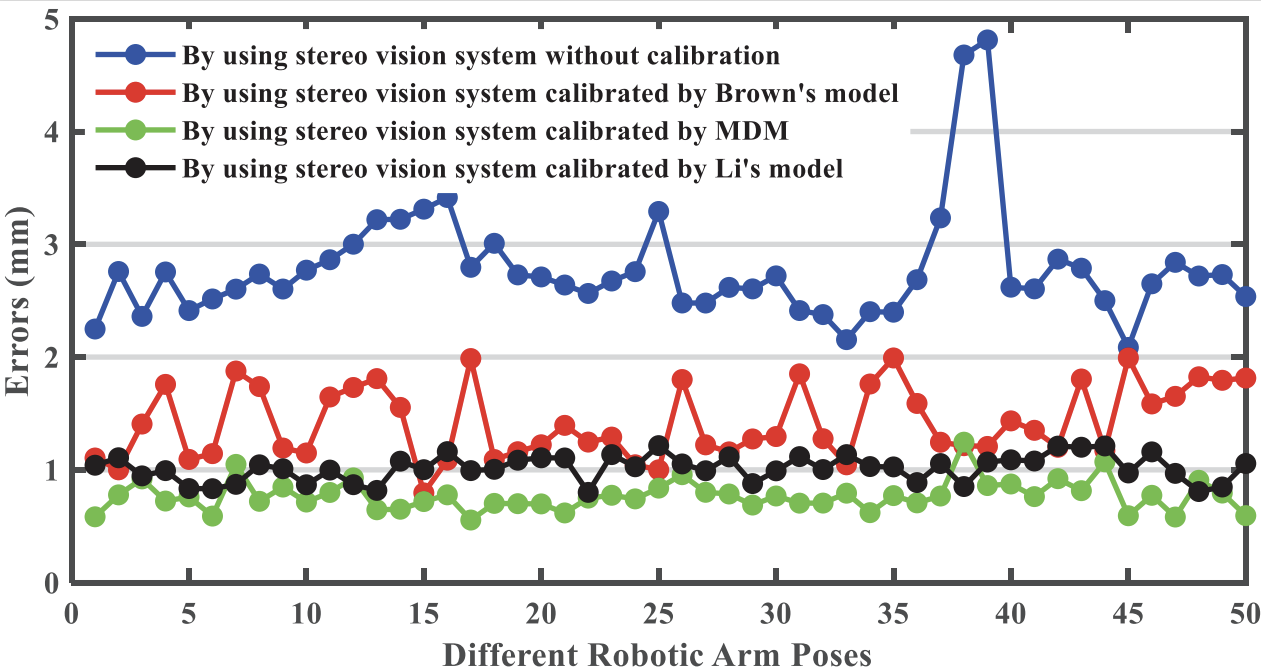}
    \caption{Robotic arm end-effector position errors by using stereo vision system calibrated by different distortion models}
    \label{fig:exp2}
\end{figure}

The results are shown in Fig. 8. The position errors were calculated by the distance between the desired position and the measurement position of the end-effector of the robotic arm. The RMS position error before the kinematic parameters calibration of the robotic arm was 2833.6 $\mu$m. By using the stereo vision system with Zhang's distortion model, MDM, and Li's distortion model for calibrating the robotic arm, the RMS position errors were 1456.7 $\mu$m, 780.6 $\mu$m, and 1020.9 $\mu$m, respectively. The above results demonstrated that the stereo vision system with higher measurement accuracy for the kinematic parameters calibration of the robotic arm will result in higher motion accuracy of the robotic arm, which will improve the performance of the robotic arm with some tasks that require high accuracy.

\section{Conclusion}
In this paper, we proposed an MDM, which considers both the radial and decentering lens distortions. Besides, a flexible three-step calibration method for the MDM was presented. The proposed flexible calibration method requires the cameras to observe a planar pattern in different orientations. Unlike the existing calibration methods for depth-dependent distortion models, the lenses are not required to be perpendicular to the planar pattern during our calibration procedures. In the experimental section, we demonstrated that the MDM improved the calibration accuracy by 56.55\% and 74.15\% compared with Li's distortion model and the traditional Brown's distortion model, respectively. Besides, the proposed calibration method was more efficient compared with Li's calibration method. In addition, the results demonstrated that the measurement accuracy using the iteration-based reconstruction method was improved by 9.08\%, on average, compared with using the non-iteration reconstruction method. In the future, we will apply our proposed stereo vision system to various scenarios, such as registration between different systems.

\vspace{-2 mm}
\end{document}